\documentclass[runningheads]{llncs}

\usepackage{eccv}

\usepackage{eccvabbrv}

\usepackage{graphicx}
\usepackage{booktabs}

\usepackage[accsupp]{axessibility}  

\usepackage[width=120mm,left=13mm,paperwidth=146mm,height=193mm,top=12mm,paperheight=217mm]{geometry}

\usepackage{tcolorbox}
\definecolor{hrefblue}{rgb}{0.12,0.49,0.85}
\usepackage[pagebackref,breaklinks,colorlinks,citecolor=hrefblue]{hyperref}

\usepackage{multirow}
\usepackage{makecell}
\usepackage{wrapfig}

\definecolor{DarkOrchid}{HTML}{9932CC}
\definecolor{VioletRed}{HTML}{D02090}
\definecolor{ForestGreen}{HTML}{228B22}

\newcommand\genixer{\textsc{Genixer}}

\newcommand\genixerL{\textsc{Genixer}$_{L}$}
\newcommand\genixerS{\textsc{Genixer}$_{S}$}

\newcommand\xg{\textcolor{VioletRed}{$X_G$}}
\newcommand\xs{\textcolor{ForestGreen}{$X_S$}}
\newcommand\xq{\textcolor{DarkOrchid}{$X_q$}}
\newcommand\xa{\textcolor{DarkOrchid}{$X_a$}}

\begin{document}

{\let\thefootnote\relax\footnotetext{$^\dag$Corresponding author}}

\title{\genixer{}: Empowering Multimodal Large Language Model as a Powerful Data Generator} 


\titlerunning{\genixer{}: An Automatic Instruction Tuning Data
Generation Pipeline}

\author{Henry Hengyuan Zhao$^1$\and
Pan Zhou$^{2\dagger}$ \and
Mike Zheng Shou$^{1\dagger}$}

\authorrunning{Henry Hengyuan Zhao et al.}

\institute{Show Lab, National University of Singapore, Singapore \and Singapore Management University, Singapore}

\maketitle

\begin{abstract}

Multimodal Large Language Models (MLLMs) demonstrate exceptional problem-solving capabilities, but few research studies aim to gauge the ability to generate visual instruction tuning data. This paper proposes to explore the potential of empowering MLLMs to generate data independently without relying on GPT-4. We introduce \genixer{}, a comprehensive data generation pipeline consisting of four key steps: (i) instruction data collection, (ii) instruction template design, (iii) empowering MLLMs, and (iv) data generation and filtering.  Additionally, we outline two modes of data generation: task-agnostic and task-specific, enabling controllable output. We demonstrate that a synthetic VQA-like dataset trained with LLaVA1.5 enhances performance on 10 out of 12 multimodal benchmarks. Additionally, the grounding MLLM Shikra, when trained with a REC-like synthetic dataset, shows improvements on 7 out of 8 REC datasets. Through experiments and synthetic data analysis, our findings are: (1) current MLLMs can serve as robust data generators without assistance from GPT-4V; (2) MLLMs trained with task-specific datasets can surpass GPT-4V in generating complex instruction tuning data; (3) synthetic datasets enhance performance across various multimodal benchmarks and help mitigate model hallucinations. The data, code, and models can be found at \href{https://github.com/zhaohengyuan1/Genixer}{https://github.com/zhaohengyuan1/Genixer}.

 
\end{abstract}

\section{Introduction}
\label{sec:intro}

Large Language Models (LLMs)~\cite{gpt3,gpt4,anil2023palm,gao2023llama} have achieved remarkable success in tackling complex natural language tasks. This progress has recently extended to Multimodal Large Language Models (MLLMs)~\cite{kosmos, kosmos2, dai2023instructblip,liu2023visual,zhang2023llama,gao2023llama,li2023otter,shikra,ye2023mplug,su2023pandagpt,zhu2023minigpt,Qwen-VL, gao2024sphinx}. These MLLMs demonstrate an exceptional ability to solve various multimodal problems, prompting the natural question: \textbf{``Are these MLLMs capable of generating multimodal data?''} This question arises due to the continued reliance on labor-intensive human labeling and the high cost of using models like GPT-4(V) for prompts. To address this gap, this paper is the first to explore the potential of current MLLMs in data generation.

\begin{figure}[t]
    \centering
    \vspace{-0.1in}
    \includegraphics[width=\linewidth]{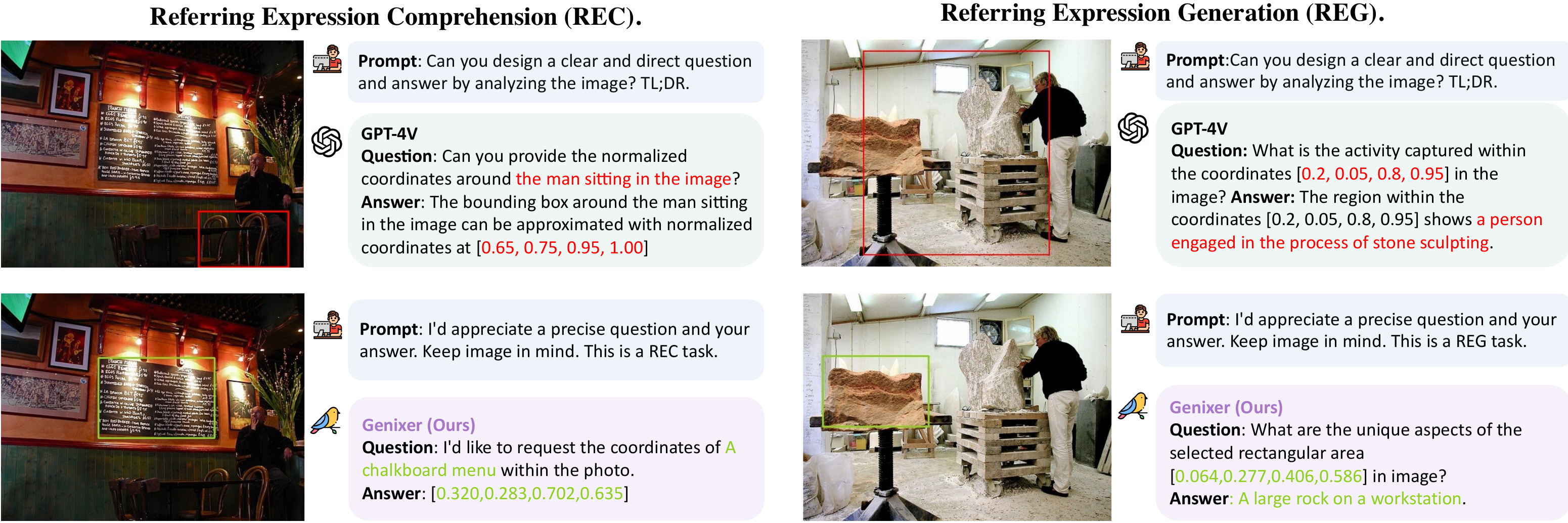}
    \caption{Two unsatisfied generation examples from GPT-4V~\cite{gpt4v}. Our proposed data generator \genixerS{} is capable of generating complex multimodal data such as REC and REG data, whereas GPT-4V fails to generate the correct bounding box.}
    \label{fig:fig1}
    \vspace{-0.2in}
\end{figure}

Recent MLLMs ~\cite{llava1.5, dai2023instructblip, chen2023minigpt2, Qwen-VL, chen2023internvl, gao2024sphinx, shikra, wang2023cogvlm} demonstrate that the visual instruction data is essential for multimodal learning. Currently, the data used for model training primarily comes from two methods: (1) reformulating current vision-language datasets into instruction-following format and (2) prompting GPT-4 to create the visual instruction datasets.
For the first way, models like InstructBLIP~\cite{dai2023instructblip} trained on traditional VL tasks, such as visual question answering (VQA) and image captioning, demonstrate effectiveness on downstream tasks. However, the datasets of these tasks suffer from a limitation in image diversity, as most of them originate from the COCO dataset~\cite{lin2014microsoft}, potentially restricting the model generalization ability. For the other way, LLaVA~\cite{llava1.5}, Shikra~\cite{shikra}, and ShareGPT4V~\cite{chen2023sharegpt4v} propose prompting GPT-4 to produce the additional visual instruction tuning data. The primary limitations of prompting GPT-4 are twofold: (1) high financial costs for large-scale dataset creation. (2) inferior performance on some complex tasks such as Referring Expression Comprehension (REC) as illustrated in Fig.~\ref{fig:fig1}. Considering these issues, we propose exploring an alternative approach to generating visual instruction tuning data by training an MLLM exclusively for data generation. The benefits of this approach are twofold: (1) it eliminates additional financial costs for data generation, and (2) it provides flexibility in producing high-quality visual instruction data for arbitrary unlabeled images.


\begin{figure}[t]
\centering
\includegraphics[width=\linewidth]{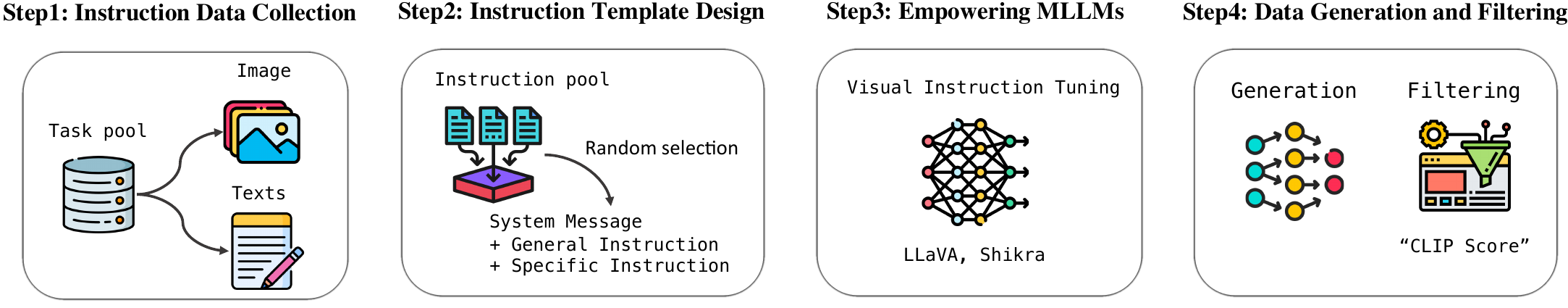}
\vspace{-0.2in}
\caption{
    The illustration of our proposed automatic data generation pipeline \genixer{}.
}
\label{fig:genxierpipeline}
\vspace{-0.2in}
\end{figure}

In this study, addressing the problem of visual instruction tuning data generation we consider two modes: task-specific visual data generation and task-agnostic data generation. Task-specific data generation is used to build synthetic datasets tailored to explicit task types, while task-agnostic data generation converts unlabeled images into various task types. To achieve this goal, we introduce a new pipeline \genixer{}, as illustrated in Fig.~\ref{fig:genxierpipeline}. This pipeline consists of four key steps: (i) instruction data collection, (ii) instruction template design, (iii) empowering MLLM, and (iv) data generation and filtering. in the first step, we focus on collecting prevalent and representative vision-language (VL) tasks as sources for data generation. The second step involves meticulously designing two-level instruction templates to enable controlled data generation for both task-specific and task-agnostic. In the third step, we select two representative MLLMs, LLaVA1.5, and Shikra, to cover two main VL task groups data generation: generic tasks and grounding tasks. Both MLLMs are trained to generate instruction data based on their possessed multimodal understanding. In the fourth step, to ensure the quality of the synthetic data, we propose removing incorrect data samples using two newly introduced automatic data filtering pipelines.

\begin{figure}[t]
\centering
\includegraphics[width=\linewidth]{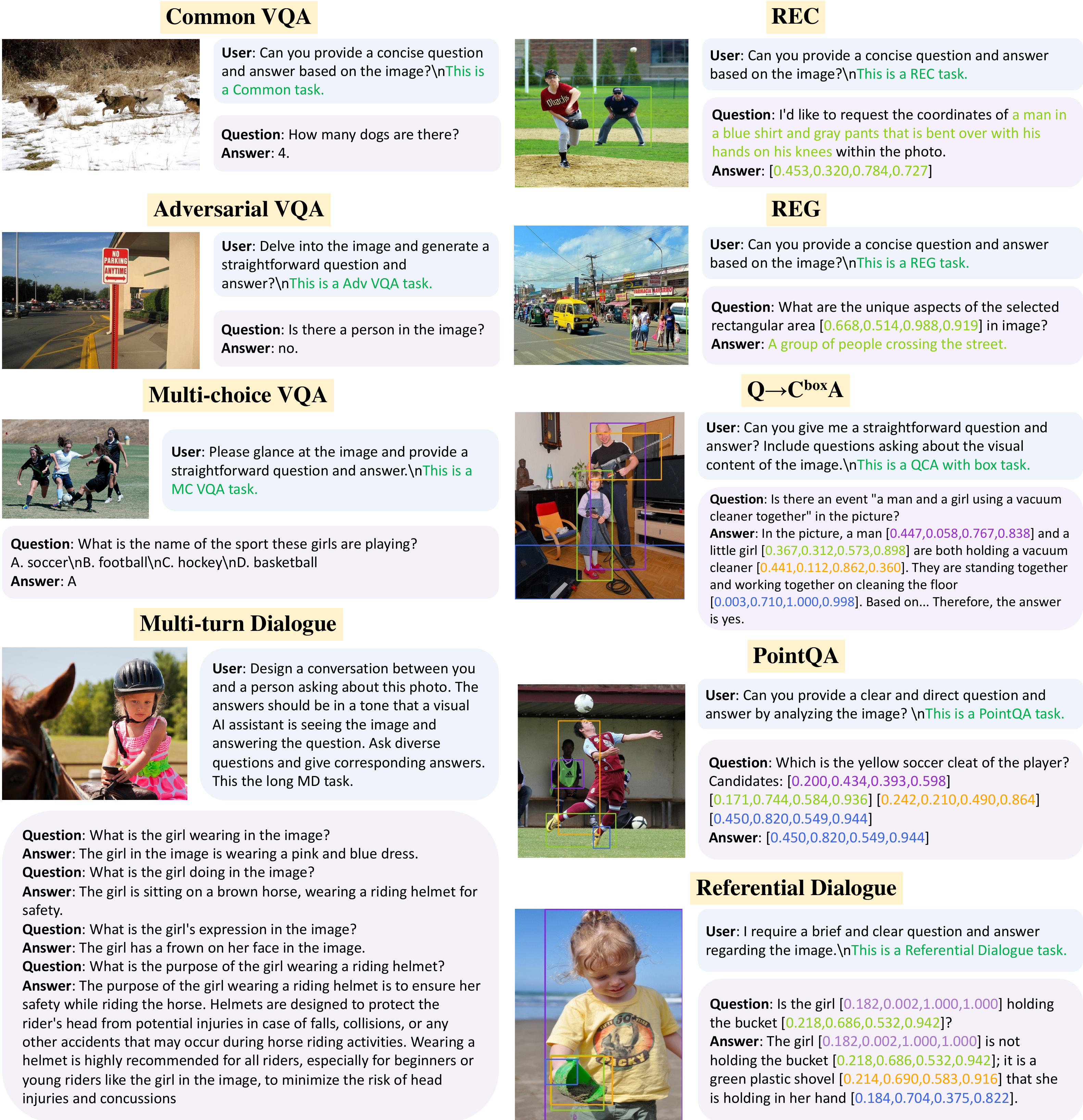}
\caption{Selected examples generated from \genixerL{} and \genixerS{}. The examples include Common VQA, Adv VQA, MC VQA, MD, and five grounding tasks.}
\label{fig:genqasamples}
\vspace{-0.2in}
\end{figure}

Using the proposed \genixer{}, we build two visual instruction tuning datasets: Genixer-915K for VQA tasks and Genixer-350K for REC tasks. Our experimental results show that training with 915K VQA-like data improves LLaVA1.5~\cite{llava1.5} on 10 out of 12 multimodal datasets and benchmarks, such as 3.8\% improvement on Vizwiz~\cite{gurari2018vizwiz}, 2.7\% on SicenceQA~\cite{lu2022learn_scienceqa}, 37.7\% on MME. Additionally, learning with 350K REC-like synthetic data, Shikra shows consistent improvements on 7 out of 8 REC datasets. Beyond quantitative evaluation, human study and dataset analysis, as detailed in Sec.~\ref{sec:statanalysis} and ~\ref{sec:humanevaluation}, demonstrate the diversity and effectiveness of generated instruction tuning data. Visualizing examples of different task types in Fig.~\ref{fig:genqasamples}, it is evident that our \genixer{} successfully produces long and meaningful answers on some challenging tasks such as Multi-turn Dialogue, PointQA, and Referential Dialogue.

Overall, our approach yields three main findings: (1) MLLMs trained using our pipeline can produce visual instruction data of comparable quality without the assistance of GPT-4V. (2) MLLMs trained with our pipeline excel at generating more complex instruction tuning data compared to GPT-4V. (3) Synthetic datasets significantly enhance MLLMs' performance on various multimodal benchmarks and help mitigate hallucinations in the models.
In summary, our contributions are as follows:
\begin{itemize}
    \item We present a holistic data generation pipeline, \textbf{\genixer{}}, capable of generating diverse visual instruction tuning data from unlabeled images. 
    \item We contribute two open-source data generator models \textbf{\genixerL{}} and \textbf{\genixerS{}} for advancing data creation in multimodal domains.
    \item We contribute two high-quality multimodal datasets, \textbf{Genixer-915K} and \textbf{Genixer-350K}, for enhancing other MLLMs in multiple benchmarks.
\end{itemize}

\section{Related Work}

\noindent\textbf{Multimodal Large Language Models.} 
Large Language Models (LLMs) showcased the remarkable complicated reasoning abilities. Some well-known open-souced LLMs include FlanT5~\cite{chung2022scaling}, OPT~\cite{zhang2022opt}, LLaMA~\cite{touvron2023llama}, Vicuna~\cite{vicuna2023} and LLaMA-2~\cite{llama2} show exceptional reasoning ability of solving math, codes problems. By leveraging these LLMs, a surge of multimodal modes~\cite{blip2,liu2023visual,dai2023instructblip,Qwen-VL,shikra,wang2023cogvlm,zhu2023minigpt,chen2023minigpt2,chen2023position,luo2023cheap,wang2023visionllm,kosmos,kosmos2,moon2023anymal,ye2023mplug,gao2023llama,lu2024deepseek,he2024bunny, lai2023lisa, cha2023honeybee,lu2023uio2} are proposed to integrate the visual information for diverse multimodal reasoning tasks such as image captioning~\cite{cococaption, agrawal2019nocaps,young2014image_flickr30k} and visual question answering~\cite{VQAv2,marino2019ok_okvqa,gqa,gurari2018vizwiz}. LLaVA~\cite{liu2023visual} is a pioneering approach that adopts a single linear layer to project the visual features extracted by CLIP~\cite{clip} encoder to the input of LLM. Different from LLaVA, InstructBLIP~\cite{dai2023instructblip} employs an instruction-aware feature extractor and obtains advanced performance on various tasks building upon the pretrained BLIP2~\cite{blip2}. Besides focusing on traditional multimodal tasks, some studies~\cite{you2024ferret, shikra, chen2023position, xu2023pixel, hanoona2023GLaMM, ren2023pixellm} focus on solving the grounding tasks with the power of LLMs. Shikra~\cite{shikra} and Ferret~\cite{chen2023position} pay attention to curating the visual instruction data, and PVIT~\cite{chen2023position} employs the region-based vision encoder. Except for these MLLMs, CogVLM~\cite{wang2023cogvlm}, Qwen-VL~\cite{Qwen-VL}, and Kosmos-2~\cite{kosmos2} explore adopting a billion-scale pertaining data corpus to enhance the model generalization ability and robustness.

\noindent\textbf{Multimodal Instruction Data.} 
High-quality visual instruction data is crucial for training an MLLM. Two main setups are as follows: 1) Transforming the current vision-language datasets into the instruction tuning format, e.g., InstructBLIP. Such choice is limited by the diversity of image sources.
2) Some approaches LLaVA, VisionLLM\cite{wang2023visionllm}, and Shikra, resort to prompting the GPT-4~\cite{gpt4} language model to generate corresponding instruction data. This way requires the image datasets to have enough captions or region-based annotations (e.g., bounding boxes), which heavily restricts the data scale. Additionally, prompting commercial LLMs incurs high costs, and even GPT-4V~\cite{gpt4v} may not address the data generation effectively on some specific tasks, as illustrated in Fig.~\ref{fig:fig1}. Recently, some works~\cite{wang-etal-2023-self-instruct, xu2024wizardlm} in natural language processing domains propose to generate text instruction tuning datasets for performance improvements. In the multimodal domain, some works have also been proposed to address image caption data generation~\cite{chen2023sharegpt4v} and text-centric instruction tuning data generation ~\cite{tang2024textsquare}. However, both approaches rely on prompting GPT-4V or Gemini Pro to build their synthetic datasets. Unlike these works, we introduce \genixer{}, an innovative pipeline that explores the capabilities of MLLMs to generate high-quality visual instruction data without assistance from current commercial LLMs. Our approach is the first to demonstrate the effectiveness of training current MLLMs to generate task-specific visual instructing tuning data and improve the MLLMs on several multimodal datasets and benchmarks.

\section{\genixer{}: An Automatic Visual Instruction Tuning Data Generation Pipeline}
\label{sec:genixermethod}

\begin{table}[t]
\centering
\caption{The statistic of tasks and datasets for training \genixer{}. We categorize the VL tasks into two categories: generic tasks and grounding tasks. Counting110K$^{\dagger}$ is built by ourselves derived from PointQA~\cite{pointqa}. POPE$^{\ddagger}$ refers to the object hallucination dataset generated by ourselves via the pipeline provided in POPE~\cite{pope}.}
\renewcommand\arraystretch{1.1}
\tabcolsep=1.4mm
\resizebox{0.76\linewidth}{!}{
\begin{tabular}{lllll}
\toprule
\textbf{Category} & \textbf{Task} & \textbf{Dataset} & \textbf{Size}\\
\midrule
\multirow{4}{*}{\rotatebox[origin=c]{90}{\textbf{Generic}}}&Common VQA & VQAv2, GQA, Counting110K$^{\dagger}$, POPE$^{\ddagger}$ & 583K \\
&Adv VQA & POPE$^{\ddagger}$& 20K \\
&MC VQA & A-OKVQA & 17K \\

&MD & VQAv2, LLaVA-Conv-58K & 108K \\
\midrule
\multirow{5}{*}{\rotatebox[origin=c]{90}{\textbf{Grounding}}} & REC & VG, RefCOCO & 1M \\
&REG & \makecell[l]{VG, RefCOCO} & 1M \\
&PointQA & \makecell[l]{PointQA Local, Visual7W} & 218K \\
&Q$\rightarrow$C$^{Box}$A & \makecell[l]{Shikra (GPT-4 Generated)} & 4K \\
&RD & \makecell[l]{Shikra (GPT-4 Generated)} & 1.8K \\
\bottomrule
\end{tabular}
}
\label{tab:genixerdatasets}
\vspace{-0.1in}
\end{table}

Though current MLLMs ~\cite{llava1.5, Qwen-VL, wang2023cogvlm} show exceptional capability of handling various multimodal tasks, rare works concentrate on visual instruction tuning data generation. To this end, We propose \genixer{}, as illustrated in Fig.~\ref{fig:genxierpipeline}, which is a novel pipeline that contains four key steps, including 1) instruction data collection, 2) instruction template design, 3) empowering MLLMs, and 4) data generation and filtering. In the following, we will elaborate on these four key steps.

\subsection{Instruction Data Collection}
\label{sec:datacollect}

In accordance with the prevalence and practical relevance of real-world VL tasks, we carefully select nine representative tasks as listed in Tab.~\ref{tab:genixerdatasets}, including  Common Visual Question Answering (Common VQA), Adversarial-based VQA (Adv VQA), Multi-choice VQA (MC VQA), Multi-turn Dialogue (MD), Referring Expression Comprehension (REC), Referring Expression Generation (REG), PointQA, Q$\rightarrow$C$^{Box}$A and Referential Dialogue (RD). We divide these tasks into two main categories, \textbf{generic tasks} and 
\textbf{grounding tasks}.

\subsection{Instruction Template Design}
\label{sec:instruction-design}

In an automatic data generation context, where image content is agnostic, preemptively determining the specific task type becomes particularly daunting, especially when it involves large-scale data creation purposes. 
Hence, we consider two key modes for visual instruction data generation: 1) task-agnostic data generation and 2) task-specific data generation.

\noindent\textbf{Two-level Instructions.} We propose a two-level instruction template for controlling the task type of generated visual instruction tuning data. The instruction template is as follows:

\begin{tcolorbox}
	[boxrule=0.5pt,boxsep=0pt,sharp corners,colback=white,colframe=black!70!white]
	\texttt{<s>}
	\text{SYSTEM MESSAGE.}
	USER: \texttt{<image>} \textcolor{VioletRed}{Generic Instruction.} \textcolor{ForestGreen}{Specific Instruction.} ASSISTANT: \textcolor{DarkOrchid}{Question: \texttt{<question>} Answer: \texttt{<answer>}} \texttt{</s>}
\end{tcolorbox}

The tags \texttt{<image>}, \texttt{<question>}, and \texttt{<answer>} serve as placeholders for inserting the tokens of the image, question, and answer, respectively. \textcolor{DarkOrchid}{Question: \texttt{<question>} Answer: \texttt{<answer>}} is the model response that needs to be predicted in a left-to-right text generation manner.

Regarding \textcolor{VioletRed}{Generic Instruction}, it allows the model to generate arbitrary types of instruction tuning data referred to as mode 1. During training, we randomly select one of 58 handwritten instructions each time. For instance,  \textit{\textcolor{VioletRed}{``Please provide a clear and direct question and answer after examining the image''}}.  Then, for \textcolor{ForestGreen}{Specific Instruction}, it is designed to determine the specific task type, such as \textit{\textcolor{ForestGreen}{``This is a Common VQA task''}}, allowing us to control the task type referred to as mode 2.

\begin{figure}[t]
\vspace{-0.1in}
\centering
\includegraphics[width=0.88\linewidth]{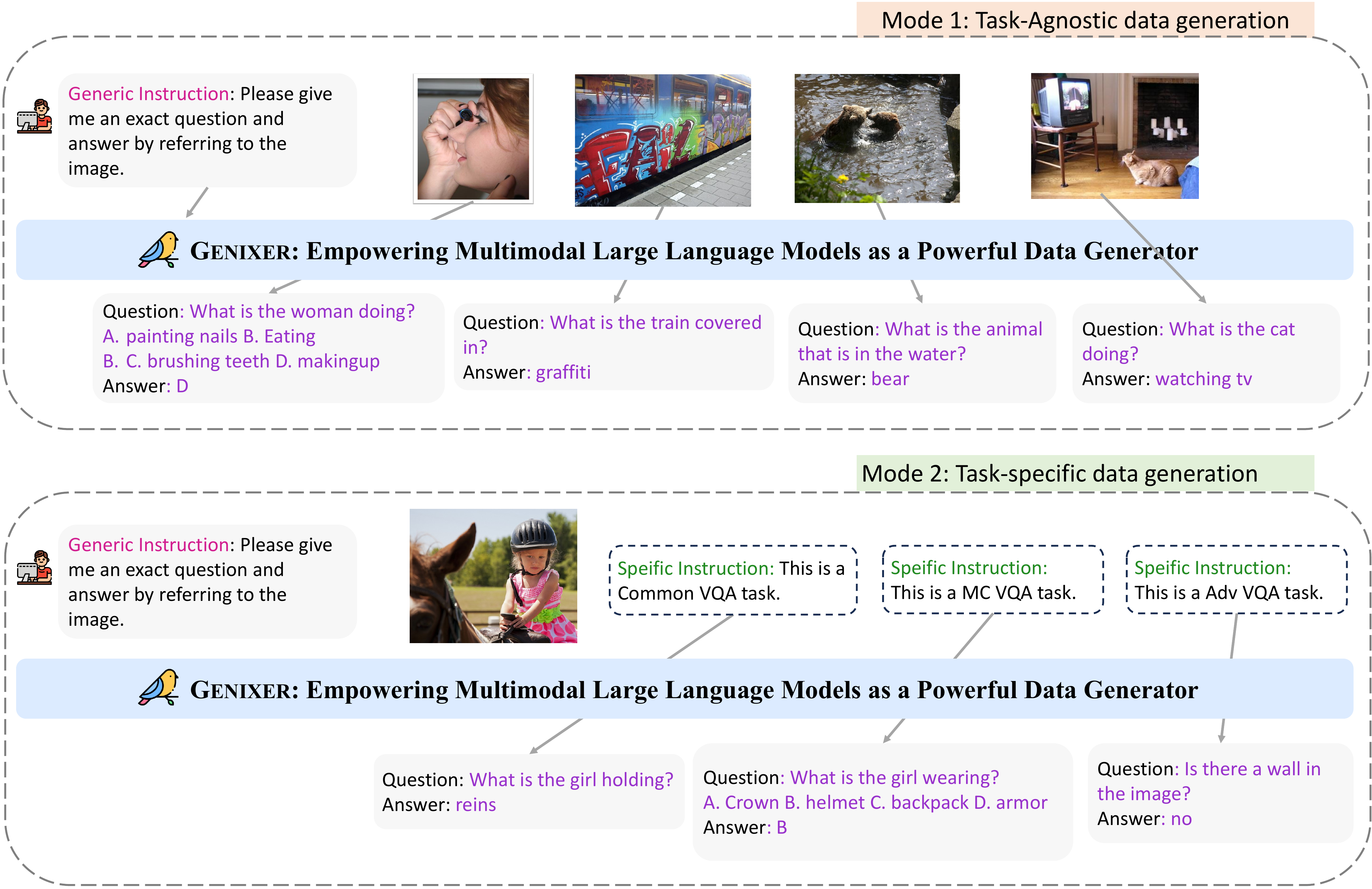}
\caption{A demonstration of two proposed instruction modes during the inference phase.}
\label{fig:instructionmode}
\vspace{-0.2in}
\end{figure}

\noindent\textbf{Controlling Constant.}
During the training phase, we set a constant $\tau$ for controlling the ratio of training samples that are exclusive with \textcolor{VioletRed}{Generic Instruction}. Consequently, in the inference phase, the model is able to switch the mode by adding specific instructions or not. For example, as illustrated in Fig.~\ref{fig:instructionmode}, the model is capable of generating various types of data in the absence of specific instructions. Simultaneously, it can produce specific outputs when guided by a detailed prompt, like \textit{``This is an MC VQA task''}.

\subsection{Empowering Current MLLMs}
\label{sec:empowermllm}

To train an MLLM with the ability of data generation, we leverage the two remarkable MLLMs, LLaVA1.5 and Shikra, as the backbone models for generic and grounding task generation. Consequently, we obtain two data generators, \genixerL{} and \genixerS{}.

\noindent\textbf{Overall Framework of \textbf{\genixerL{}} and \textbf{\genixerS{}}.} For brevity, we denote  the MLLM model as $F_M$, the generic and specific instructions as \xg{} and \xs. Then, given an image $X_I$, training objective is to make $F_M$ to generate question \xq{} and corresponding answer \xa{}:
\begin{equation}
   \textcolor{DarkOrchid}{X_q}, \textcolor{DarkOrchid}{X_a} = F_M(\textcolor{VioletRed}{X_G}, \textcolor{ForestGreen}{X_S}, X_I).
\end{equation}
To this end, we follow previous MLLMs\cite{llava1.5, dai2023instructblip}, and design the training objective in an autoregressive manner:
\begin{equation}
    \max \sum_{i=i}^{L} \log p(X_{o}|(\textcolor{VioletRed}{X_G},\textcolor{ForestGreen}{X_S},X_I) = \prod_{i}^{L}p_{\theta}(x_i|(\textcolor{VioletRed}{X_G}, \textcolor{ForestGreen}{X_S}, X_I, X_{o,<i}),
\end{equation}
where $X_o$ is the whole sentence compose \xq{} and \xa{}, and $x_i$ is current prediction token. $L$ is the length of the model response sequence. $\theta$ denotes the total trainable parameters in  $F_M$ (e.g., the parameters of projector and LLM with LLaVA1.5 backbone).

\noindent\textbf{Training of \genixerL{}.} \genixerL{} trains the pretrained LLaVA1.5~\cite{llava1.5} for four kinds of generic tasks, including  Common VQA, Adv VQA, MC VQA, and MD. As summarized in Tab.~\ref{tab:genixerdatasets}, we only sample a subset of original datasets from these task types for training efficiency. The controlling constant $\tau$ is set to 0.2, 0.2, 0.5, and 0.2, respectively. The different ratios are because of the different data sizes of these task types. These values are chosen manually to keep the training data balanced. Finally, we use AdamW~\cite{adamw} optimizer with a learning rate of $1 \times 10^{-5}$ and a batch size of 128 for one epoch training takes about 14 hours.

\noindent\textbf{Training of \genixerS{}.} We utilize the Shikra as the backbone model to train an MLLM with the ability to generate grounding task data. As shown in Tab.~\ref{tab:genixerdatasets}, the dataset size for RD is relatively small. To counteract potential biases in the generation, \genixerS{} that finetunes a SoTA model Shikra~\cite{shikra} on region-based tasks adopts two-phased training. The first phase focuses on the REC and REG data generation. In the second phase, \genixerS{} adds PointQA, Q$\rightarrow$C$^{Box}$A, and RD data while deliberately reducing REC and REG data for data trade-off. To train the \genixerS{}, we set the $\tau$ equal to 0.5 for each task type. We utilize the AdamW optimizer, applying learning rates of $3 \times 10^{-5}$ for the first phase and $1 \times 10^{-5}$ for the second phase. The batch sizes are set to 128 and 64, respectively.

\begin{figure}[t]
    \centering
    \vspace{-0.1in}
    \includegraphics[width=0.86\linewidth]{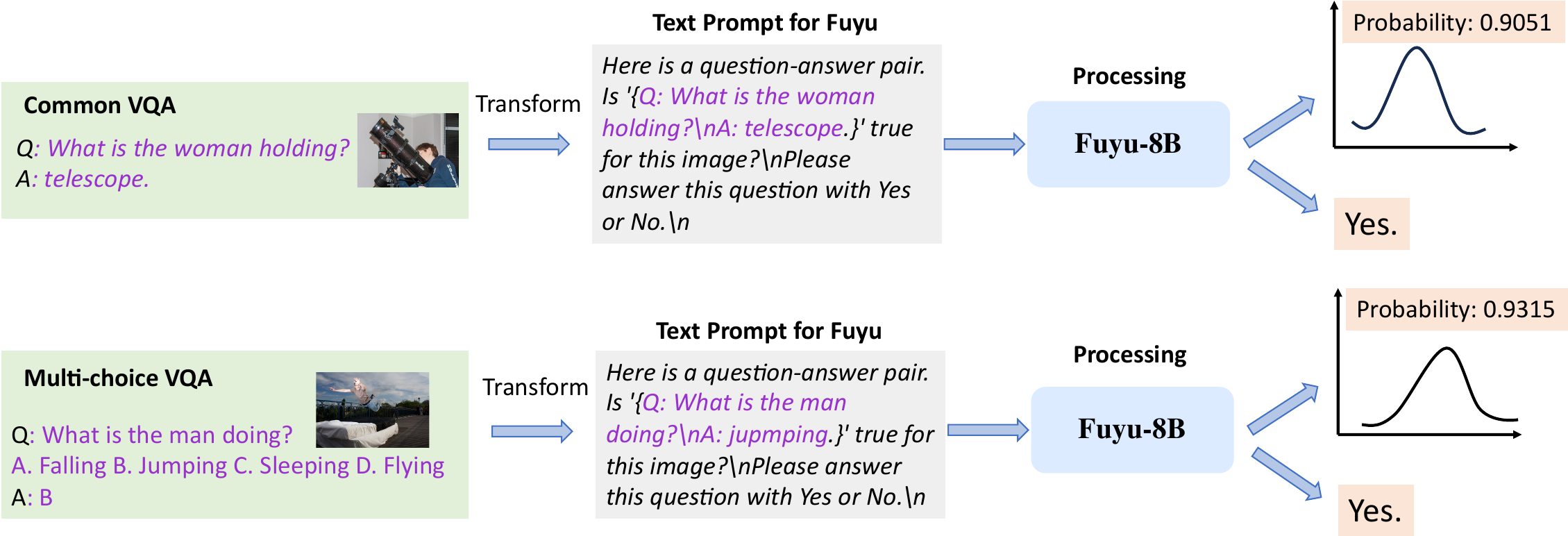}
    \caption{The illustration of proposed Fuyu-driven data filtering framework.
    The outputs of the framework compose a probability and a direct answer.}
    \label{fig:fuyufilter}
    \vspace{-0.2in}
\end{figure}

\subsection{Automatic Data Generation and Filtering}
\label{sec:dgandfilter}
Here, we introduce the image sources used for data generation. Additionally, recognizing that there are no ready-made state-of-the-art models available for the VQA data filtering, we introduce two data filtering pipelines to remove the incorrect data samples, which is essential for subsequent model training.

\noindent\textbf{Data Generation and Filtering of \genixerL{} on The Generic Tasks.}  
We utilize the mixed dataset comprising 558K images from the LAION~\cite{laion5b}, CC3M~\cite{cc3m}, and SBU~\cite{sbu}, as described in \cite{liu2023visual}, and further adopt the 830K images from the original SBU datasets. We directly feed these 1.4M images to the trained \genixerL{} and then designate the specific instruction \xs{} = \textit{``This is the Common VQA task''}, and finally we produce the 1.4M raw data samples.

To assess the quality of generic task data, we design a Fuyu-driven data filtering framework to automatically filter the samples, which may include incorrect questions or incorrect answers. We design the text prompt as follows:
\begin{tcolorbox}
	[boxrule=0.5pt,boxsep=0pt,sharp corners,colback=white,colframe=black!70!white]
	Here is a question-answer pair. Is $\{$Q:\xq{}$\backslash$nA:\xa{}$\}$ true for this image?$\backslash$n Please answer this question with Yes or No.$\backslash$n
\end{tcolorbox}

For filtering Common VQA and Adv VQA tasks, we substitute the variables \xq{} and \xa{} with the generated questions and answers, respectively. In the case of MC VQA evaluation, we replace the option letter (e.g., ``B'') with the corresponding option content (e.g., ``Jumping'') and then convert the format to match that of Common VQA for processing by Fuyu-8B~\cite{fuyu-8b}, as shown in Fig. \ref{fig:fuyufilter}. For the MD task, we decompose multi-turn dialogue into individual single-turn instances for filtering.

Rather than prompt Fuyu-8B to directly output ``Yes'' or ``No'' as the filtering label, we calculate the probability of predicting the ``Yes'' as follows:
\begin{equation}
	\label{eqn:prob}
	P(Y_r|X_I, \textcolor{DarkOrchid}{X_q}, \textcolor{DarkOrchid}{X_a}) = \prod_{i}^{L} p(y_i|X_I, \textcolor{DarkOrchid}{X_q}, \textcolor{DarkOrchid}{X_{a,<i}}),
\end{equation}
where $Y_{r}$ is the predicted response and $L$ is the length the total response sequence. Then, we propose a threshold $\lambda$ to control the filtering in the following manner:

\begin{equation}
	S^n = \left\{
	\begin{array}{ll}
		\text{True}, & \text{if $Y_r$ = \text{Yes} and $P(Y^n_r) > \lambda$} \\
		\text{False}, & \text{if $Y_r$ = Yes and $P(Y^n_r) \leq \lambda$} \\
		\text{False}, & \text{if $Y_r$ = No},
	\end{array}
	\right.
\end{equation}
where $S^n$ is the filter label representing keeping or removing the current sample. $P(Y^n_r)$ denotes the probability of the result ``Yes'' of $n$-th candidates.  By setting $\lambda = 0.7$, we filter the 1.4M raw VQA triplets to 915K instances. We name this VQA-like synthetic dataset \textbf{Genixer-915K}.

\noindent\textbf{Data Generation and Filtering of \genixerS{}  on The Grounding Tasks.}  
\genixerS{} adopts the same image resources in generic tasks for generating grounding-based instruction tuning data. After feeding the 1.4M image corpus to \genixerS{} by set the data type as REC, we get 1.4M raw data.

To assess the quality of these REC data, we propose a CLIP-driven data filtering framework. Specifically, we first use the regular expression to extract the text expression and region coordinates from the raw generated sentence and then conduct the following three steps to filter the generated data in a coarse-to-fine manner.
1) Removing the wrong question or answer formats (e.g., wrong coordinate format). 2) Removing the bonding box whose width or height is smaller than 50. 3) Employing OpenCLIP-L~\cite{openclip} model for calculating the similarity score between the text expression and their corresponding image region, discarding samples with CLIP scores below 0.6. For example, consider one REC sample with the Question \textit{``I need the coordinates of the person at the bottom left of the image. Can you assist?''} and the Answer \textit{``[0.005,0.332,
0.249,0.984]''}. Here, \textit{``person at the bottom left of the image''} is the text expression, and the coordinate of the referring region is \textit{``[0.005,0.332,0.249,0.984]''}. 

By applying a threshold of 0.6, we filter out 350K instances from 1.4M images and name this REC-like synthetic dataset as \textbf{Genixer-350K}.

\section{Experiments}
In this section, we evaluate the quality of synthetic datasets from several aspects, including statistical analysis, human evaluation, evaluation via training MLLMs, ablation study, visualizations, and user study.

\subsection{Statistical Analysis}
\label{sec:statanalysis}

\begin{figure}[t]
	\centering
	\begin{minipage}[c]{\linewidth}
		\begin{subfigure}[b]{0.2\linewidth}
			\includegraphics[width=\linewidth]{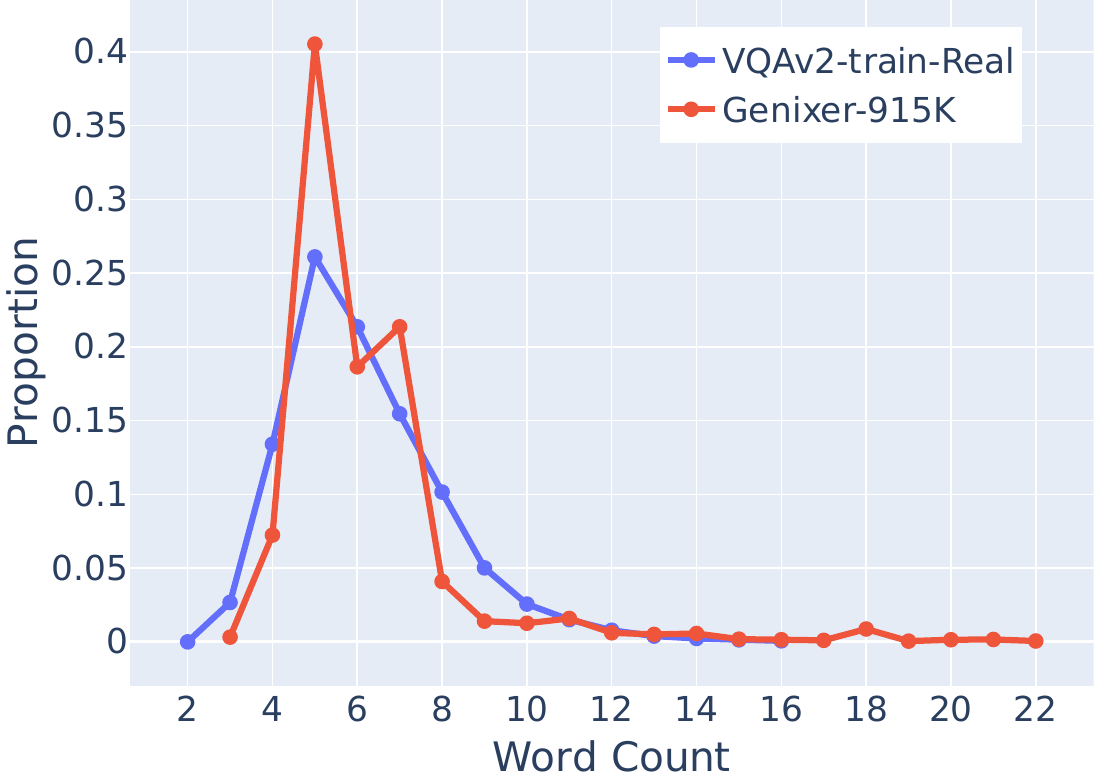}
			\caption{Question length Comparison.}
		\end{subfigure}
		\hfill
            \begin{subfigure}[b]{0.38\linewidth}
			\includegraphics[width=\linewidth]{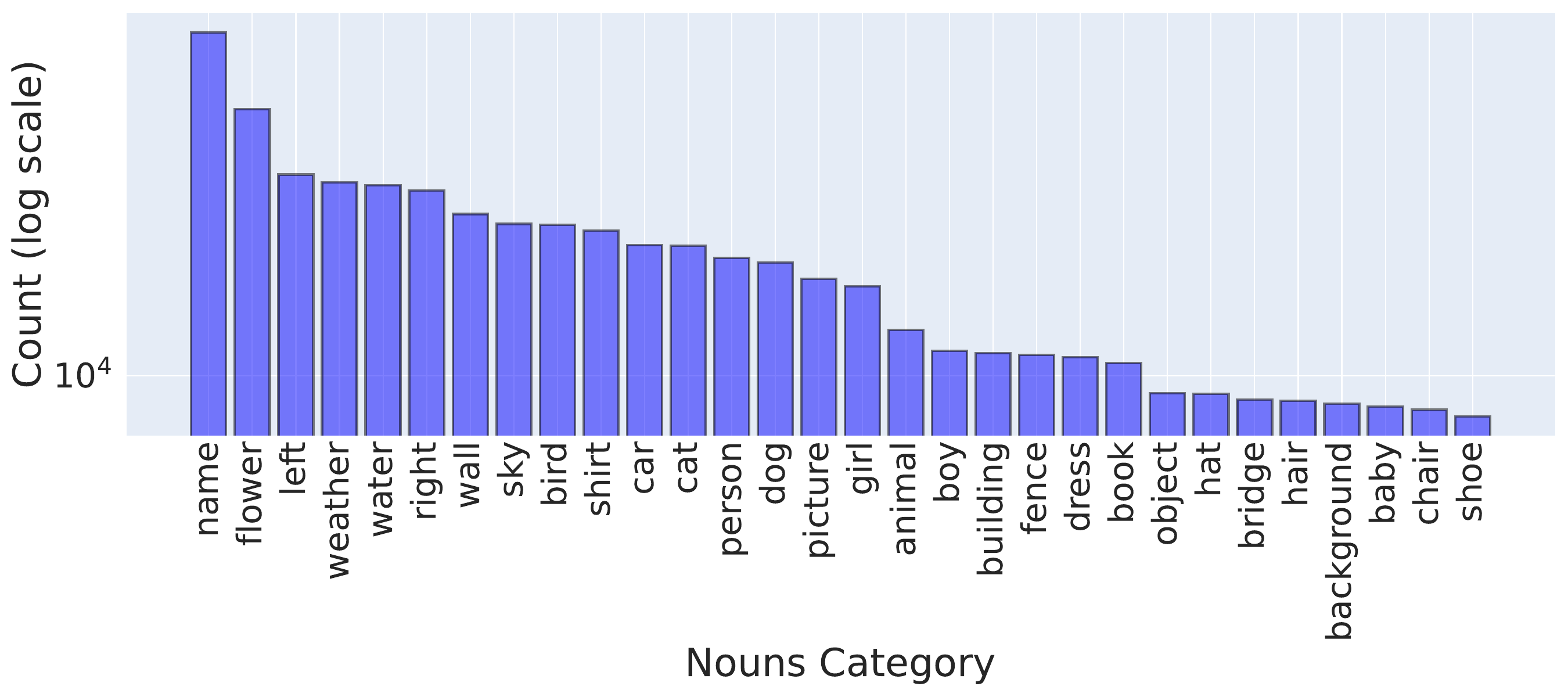}
			\caption{Nouns distribution.}
            \end{subfigure}
            \hfill
            \begin{subfigure}[b]{0.38\linewidth}
			\includegraphics[width=\linewidth]{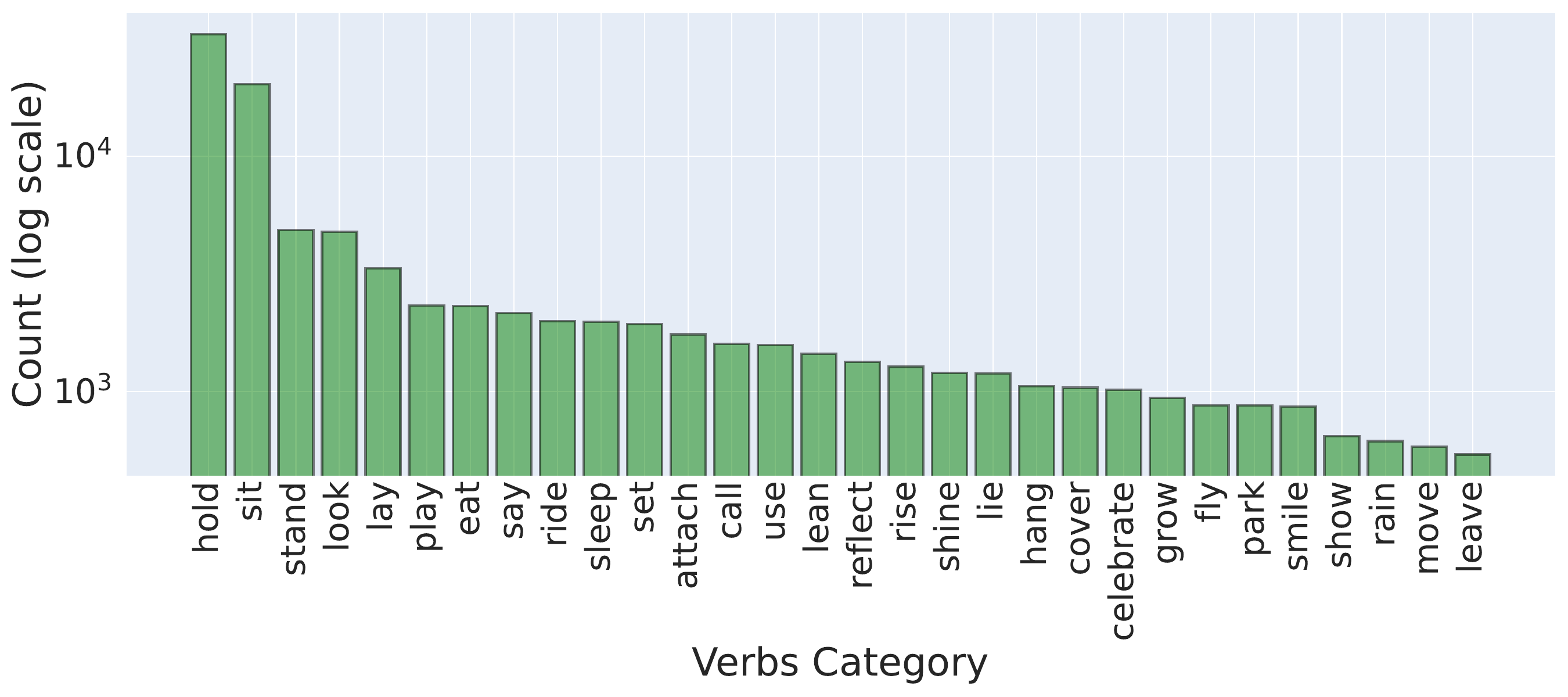}
			\caption{Verbs distribution.}
		\end{subfigure}
	\end{minipage}
        \vspace{-0.1in}
	\caption{
		The statistics of VQA-like dataset Genixer-915K. 
	}
	\label{fig:generalstat}
\end{figure}

To evaluate the generation quality of \genixerL{}, we conduct a comparative analysis using the VQAv2~\cite{VQAv2} training set as a comparison. Fig.~\ref{fig:generalstat} (a) illustrates the distribution of question lengths of VQAv2 and \textbf{Genixer-915K}. Notably, our dataset exhibits a significant long tail, indicating a higher proportion of longer sentences compared to VQAv2. Additionally, the distribution of nouns and verbs, depicted in Fig.~\ref{fig:generalstat} (b) and (c), showcases the diverse vocabulary present in the generated questions. To assess data quality, we utilize Flickr30K~\cite{young2014image_flickr30k} as the image source and employ \genixerL{} to generate three representative data types, as detailed in Sec.~\ref{sec:dgandfilter}. Results in Tab.~\ref{tab:fuyuresults} demonstrate that our generated data achieves an accuracy exceeding 80\%, with the highest accuracy observed in the MC VQA category. Furthermore, our generated data exhibits a high probability (0.8) of being classified as ``Yes'' across all three data types. This is corroborated by the probability distribution depicted in Fig.~\ref{fig:genixerprob}, affirming the high quality of data produced by \genixerL{} in generating diverse instruction tuning data.

\begin{figure}[t]
	\centering
	\begin{minipage}[c]{\linewidth}
		\begin{subfigure}[b]{0.3\linewidth}
			\includegraphics[width=\linewidth]{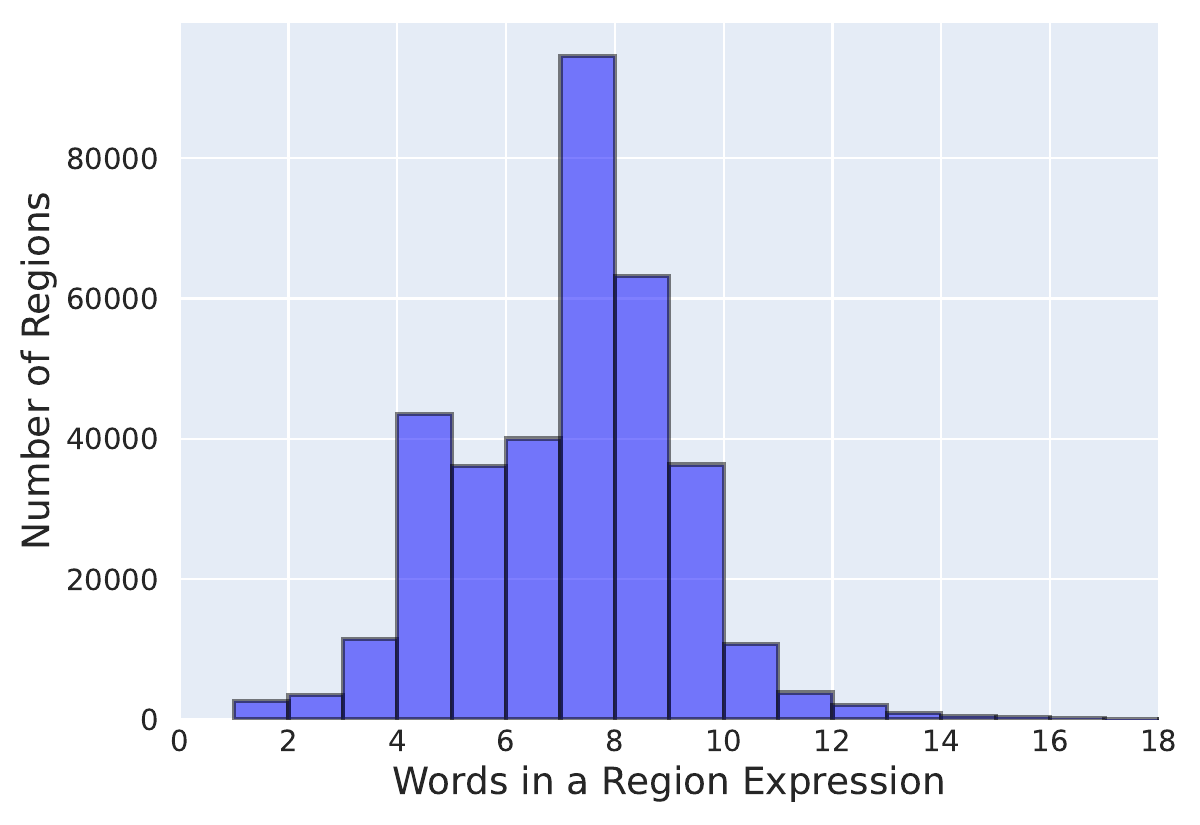}
			\caption{Expression Length.}
		\end{subfigure}
		\hfill
		\begin{subfigure}[b]{0.3\linewidth}
			\includegraphics[width=\linewidth]{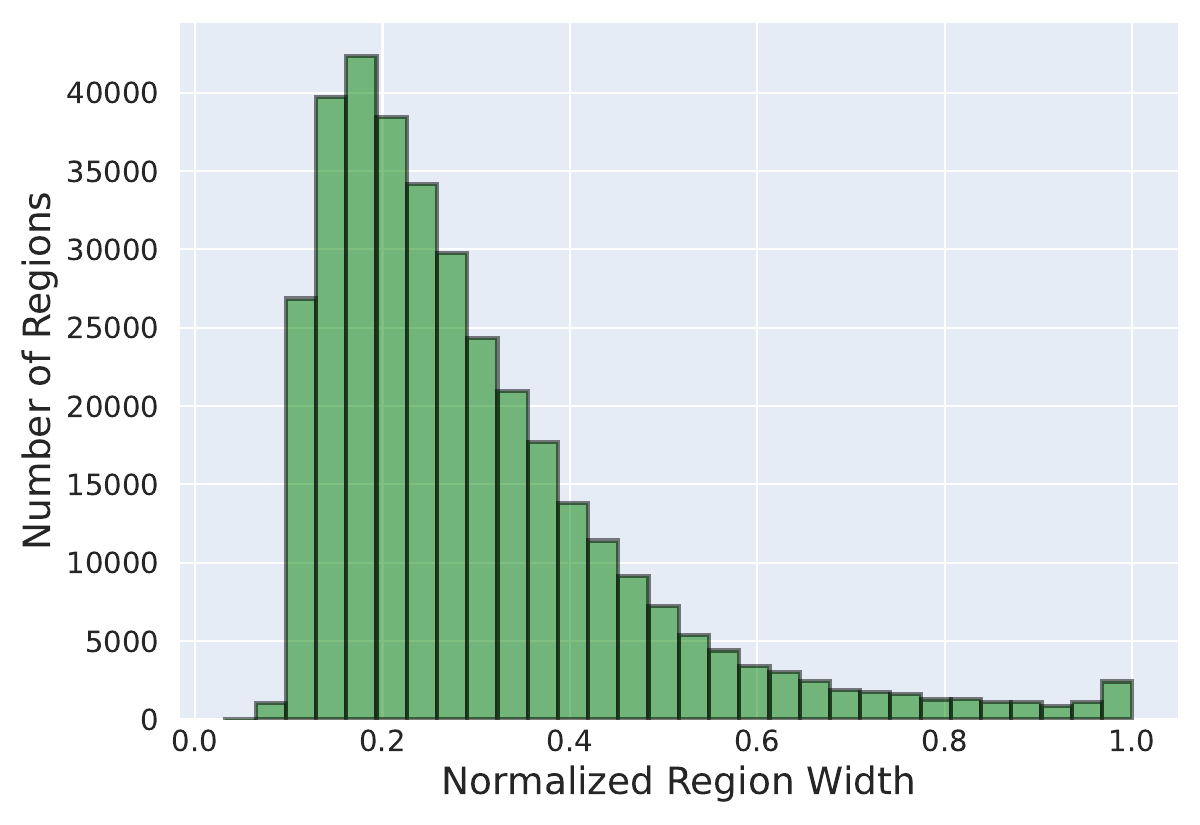}
			\caption{Region width.}
		\end{subfigure}
		\hfill
		\begin{subfigure}[b]{0.3\linewidth}
			\includegraphics[width=\linewidth]{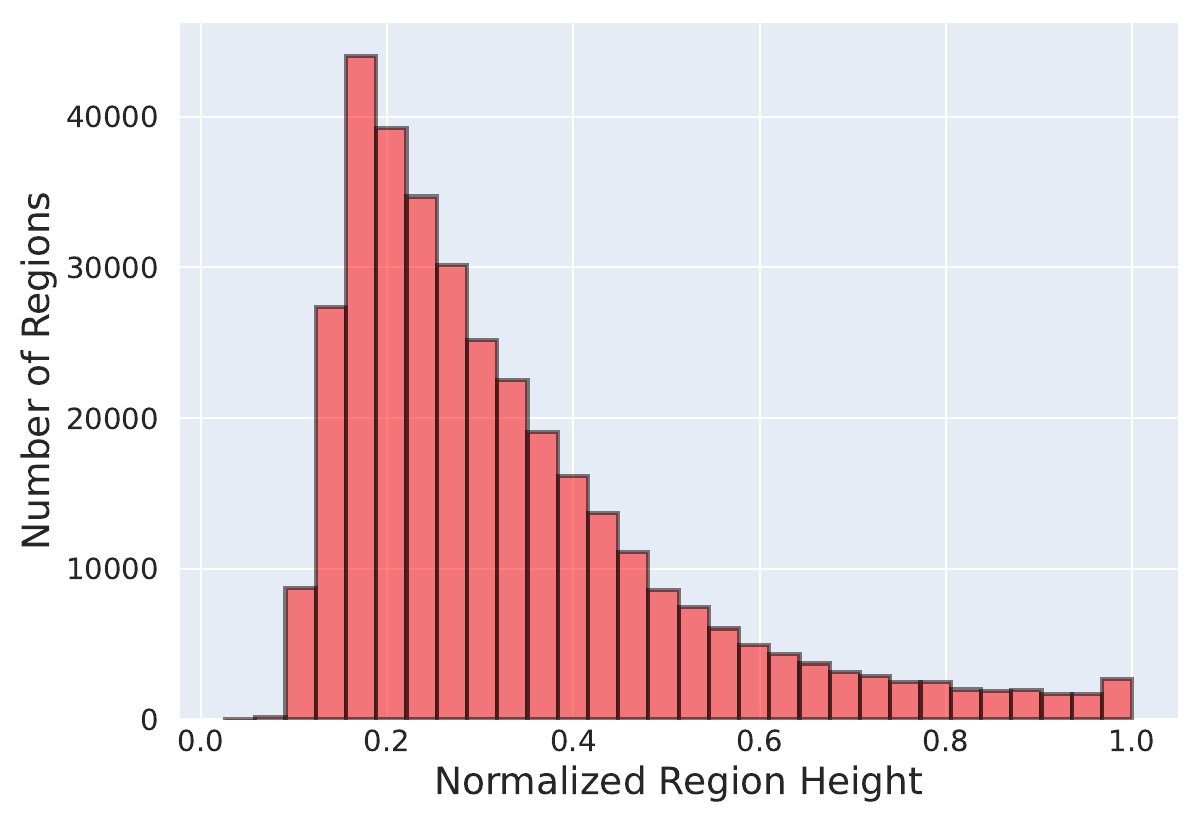}
			\caption{Region Height.}
		\end{subfigure}
  
	\end{minipage}
        \vspace{-0.1in}
	\caption{
		The statistics of REC-like dataset Genixer-350K. 
	}
	\label{fig:groudningstat}
\end{figure}

The statistics of the synthetic dataset \textbf{Genixer-350K} are presented in Fig.~\ref{fig:groudningstat}, showcasing metrics related to expression length, region width, and height. Tab.~\ref{tab:statrecdata} offers a comparative analysis of our dataset, highlighting the larger collection of images and expression lengths in Genixer-350K compared to other grounding-based datasets.

\subsection{Human Evaluation}
\label{sec:humanevaluation}
We conduct the human evaluation to manually analyze the generated question type and corresponding correctness. We employ \genixerL{} to generate QA samples without specific the data type. Since the image content can affect the generation types, we randomly chose 100 images from the COCO validation set as the held-in set and 100 images from Flickr30K as the held-out set. As shown in Tab.~\ref{tab:manualeval}, 
we divide the questions into seven types, such as ``Action'' and ``Color''. One can observe that 
``Object Type'' and ``Relative position'' are the two most popular question types of both held-in and held-out datasets. Among held-in samples, ``Action'' question also has a notable proportion with 92\% correctness. Different from the held-in set, 38 out of 100 samples belong to ``Object Type'' and 23 out of 100 ``Relative Position'' in the held-out set. In all of these types, the correctness of held-in dataset is slightly higher than held-out dataset.

\begin{figure}[t]
    \begin{minipage}[c]{0.5\linewidth}
    \captionof{table}{Fuyu-8B evaluation result on Flickr30K image dataset. Accuracy refers to the ``Yes'' prediction. Prob. represents the probability.}
    \label{tab:fuyuresults}
    \renewcommand\arraystretch{1.2}
    \centering
    \resizebox{0.9\linewidth}{!}{
        \begin{tabular}{@{}lcc@{}}
            \toprule
            Data Type &  Accuracy($\sim$\%)  & Average Prob. \\
            \midrule
            Common VQA & 82.4 & 0.8186\\
            MC VQA & 87.8 & 0.8721\\
            MD & 82.5 & 0.8252 \\
            \bottomrule
        \end{tabular}
    }
    \end{minipage}
    \hfill
    \begin{minipage}[c]{0.48\linewidth}
    \centering
    \vspace{0.1in}
    \includegraphics[width=0.7\linewidth]{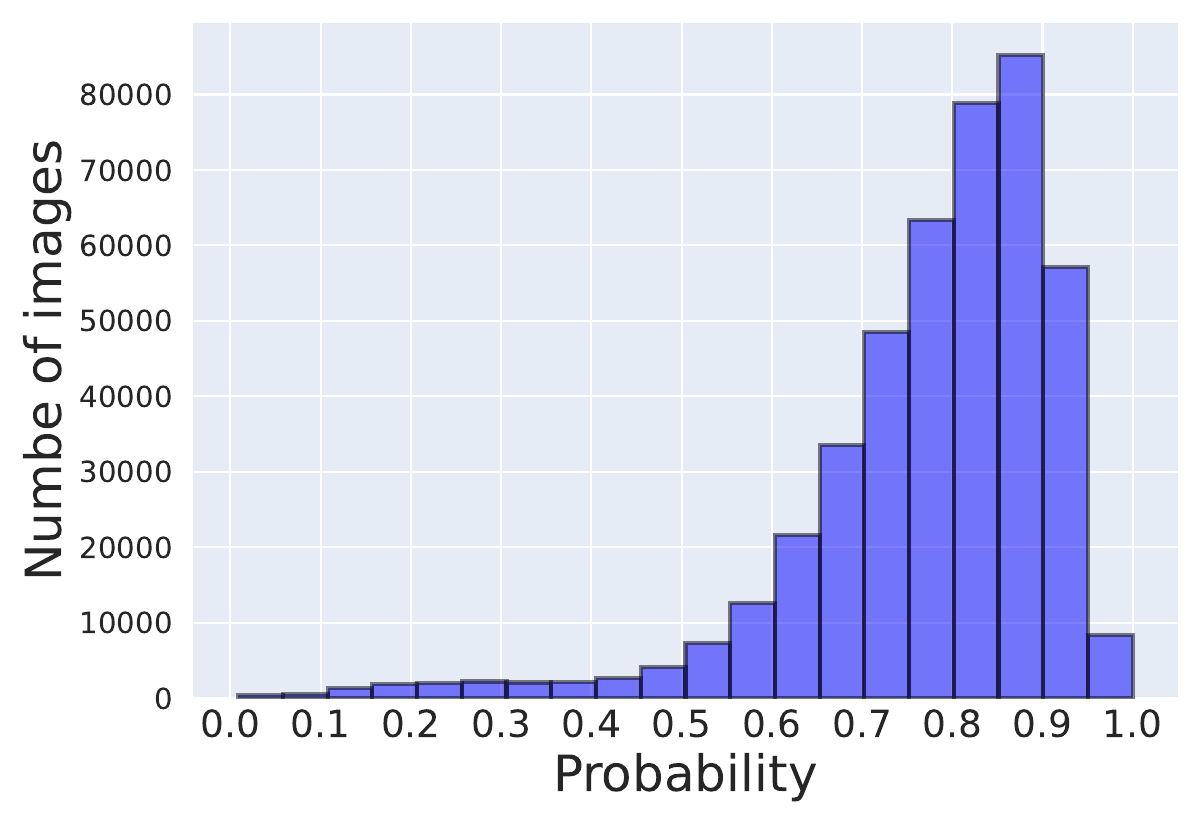}
    \vspace{-0.1in}
    \caption{The distribution of the probability by Fuyu-8B evaluation on Genixer-915K.}
    \label{fig:genixerprob}
    \end{minipage}

\end{figure}

\begin{table}[t]
    \renewcommand\arraystretch{1.1}
    \begin{minipage}[t]{0.48\linewidth}
    \caption{Comparison of images, objects, and average length between Genixer-350K with other visual grounding datasets.}
    \label{tab:statrecdata}
    \centering
    \vspace{-0.1in}
    \resizebox{0.92\linewidth}{!}{
    \begin{tabular}{lrrc}
    \toprule
        Dataset & Images & Objects & Avg. Length \\ \midrule
        Flickr Entities~\cite{flickr_entity} & 31,783 & 275,775 & -- \\
        RefCOCOg~\cite{refcocog} & 26,711 & 54,822 & 8.43 \\ 
        RefCOCO~\cite{refcoco} & 19,994 & 50,000 & 3.61 \\
        RefCOCO+~\cite{refcoco} & 19,992 & 49,856 & 3.53 \\
        Visual Genome~\cite{vg} & 108,077 & 4,102,818 & -- \\
        \midrule
        Genixer-350K & 350,000 & 447,801 & 6.67 \\
    \bottomrule
    \end{tabular}
    }
    \end{minipage}
    \hfill
    \begin{minipage}[t]{0.48\linewidth}
    \caption{The human evaluation on 100 randomly selected examples from held-in (COCO Val) and held-out (Flickr30K) datasets. }
    \vspace{-0.1in}
    \label{tab:manualeval}
    \resizebox{\linewidth}{!}{
        \begin{tabular}{@{}lcccc@{}}
            \toprule
            \multirow{2}{*}{Question Type} & \multicolumn{2}{c}{Held-in (COCO val)}& \multicolumn{2}{c}{Held-out (Flickr30K)} \\ 
            & \#Samples  & Correct ($\sim$\%) & \#Samples  & Correct($\sim$\%)  \\
            \midrule
            Action & 13 & 92 & 17 & 88\\
            Color & 8 & 75 & 4 & 75\\
            Counting & 6 & 83 & 3 & 66 \\
            Object Type & 23 & 87 & 38 & 76 \\
            Relative Position& 32 & 75 & 23 & 65\\
            Yes/No & 2 & 50 & 4 & 100 \\
            Others & 16 & 81 & 11 & 82\\
            \bottomrule
        \end{tabular}
    }
    
    \end{minipage}
\end{table}

\subsection{Evaluation via Training MLLMs}
\label{sec:trainmllm}

Unless specifically stated otherwise, all experiments were conducted on an 8 Nvidia A100 (40G) GPU setup.

\noindent\textbf{Evaluation on General Tasks.}

\noindent\textbf{Benchmarks.} Following the baseline LLaVA1.5~\cite{llava1.5}, we evaluate the enhanced model on 12 multimodal datasets and benchmarks. We select five generic VQA datasets, including VQAv2~\cite{VQAv2}, GQA~\cite{gqa}, VizWiz~\cite{gurari2018vizwiz}, ScienceQA~\cite{lu2022learn_scienceqa}, and TextVQA~\cite{textvqa}.
We test the multimodal benchmarks on MME~\cite{fu2023mme}, MMBench~\cite{liu2023mmbench}, SEED-Bench~\cite{li2023seedbench}, LLaVA-Bench~\cite{liu2023visual} and MM-Vet~\cite{yu2024mmvet}. Additionally, we also test the model on hallucination benchmark POPE~\cite{pope}.

\noindent\textbf{Main results.} As pointed out in \cite{dai2023instructblip} and \cite{llava1.5}, the ratio of dataset mixture is crucial for the model training. We proceed to add the large-scale synthetic dataset Genixer-915K into the pretraining stage rather than the finetuning stage for a fair comparison, adhering to the same training protocols utilized by LLaVA1.5. For the finetuning stage, we meticulously curated 8K VQA-like data samples, selected based on Fuyu-8B's probability range of 0.5 to 0.7 with the image resource from SBU~\cite{sbu}. The probability range was chosen based on findings that higher probabilities correlate with simpler, more straightforward VQA instances. By selecting data within this specific range, our model can learn more challenging VQA samples. From Tab.~\ref{tab:generalresults}, one can observe consistent enhancements across 10 out of 12 benchmarks compared with vanilla LLaVA1.5.  Indeed, on several tasks, our generated data can make significant improvements, e.g.,  3.8\%  on VizWiz, 2.9\%  on SciencQA, and  37.7 scores on the MME benchmark. While the improvement in the SEED benchmark is not as obvious as we initially expected, this is primarily because the synthetic data belongs to the Common VQA type while many questions of SEED require outside knowledge to solve which is out of the scope of Genixer-915K. The performance on LLaVA-Bench$^\text{W}$ and MM-Vet shows a slight decline, possibly due to their smaller scale and the utilization of GPT-4 as the evaluation metric. This could introduce additional uncertainty and potential biases into the results.

\begin{table}[t!]
\centering
\caption{Comparison with SoTA methods on 12 benchmarks. $^*$ represents the train set used in training. All abbreviated names of benchmarks are following \cite{llava1.5}. $^{\dagger}$ indicates results we reproduced since the original results could not be replicated, as mentioned by the author in the corresponding GitHub issue.}
\scalebox{0.64}{
\renewcommand\arraystretch{1.2}
\begin{tabular}{ll|ccccc|ccccccc}
\toprule
Method & LLM  & VQA$^\text{v2}$ & GQA & VizWiz & SQA$^\text{I}$ & VQA$^\text{T}$ & POPE & MME & MMB & MMB$^\text{CN}$ & SEED$^\text{I}$ & LLaVA$^\text{W}$ & MM-Vet \\
\midrule
BLIP-2~\cite{blip2} & Vicuna-13B & 41.0 & 41.3 & 19.6 & 61.0 & 42.5 & 85.3 & 1293.8 & -- & -- & 49.7 & 38.1 & 22.4 \\
InstructBLIP~\cite{dai2023instructblip} & Vicuna-7B & -- & 49.2 & 34.5 & 60.5 & 50.1 & -- & -- & 36.0 & 23.7 & 58.8 & 60.9 & 26.2 \\
InstructBLIP~\cite{dai2023instructblip} & Vicuna-13B & -- & 49.5 & 33.4 & 63.1 & 50.7 & 78.9 & 1212.8 & -- & -- & -- & 58.2 & 25.6 \\
Shikra~\cite{shikra} & Vicuna-13B & 77.4$^*$ & -- & -- & -- & -- & -- & -- & 58.8 & -- & -- & -- & -- \\
IDEFICS-9B~\cite{laurenccon2023obelics} & LLaMA-7B & 50.9 & 38.4 & 35.5 & 44.2 & 25.9 & -- & -- & 48.2 & 25.2 & 44.5 & -- & -- \\
IDEFICS-80B~\cite{laurenccon2023obelics} & LLaMA-65B & 60.0 & 45.2 & 36.0 & \underline{68.9} & 30.9 & -- & -- & 54.5 & 38.1 & 53.2 & -- & -- \\
Qwen-VL~\cite{Qwen-VL} & Qwen-7B & \underline{78.8}$^*$ & 59.3$^*$ & 35.2 & 67.1 & \textbf{63.8} & -- & -- & 38.2 & 7.4 & 62.3 & -- & -- \\
Qwen-VL-Chat~\cite{Qwen-VL} & Qwen-7B & 78.2$^*$ & 57.5$^*$ & 38.9 & 68.2 & \underline{61.5} & -- & 1487.5 & 60.6 & 56.7 & 65.4 & -- & -- \\
\midrule

LLaVA-1.5 & Vicuna-7B & 78.5$^*$ & \underline{62.0}$^*$ & \underline{50.0} & 66.8 & 58.2 & \underline{85.9} & \underline{1465.0}$^{\dagger}$ & \underline{64.3} & \underline{58.3} & \underline{66.2} & \textbf{65.4} & \textbf{31.1} \\

LLaVA-1.5$_{+\text{G-910K(ours)}}$ & Vicuna-7B & \textbf{79.1}$^*$ & \textbf{63.1}$^*$ & \textbf{53.8} & \textbf{69.7} & 59.0 & \textbf{87.3} & \textbf{1502.7} & \textbf{65.3} & \textbf{59.4} & \textbf{66.6} & \underline{64.0} & \underline{30.1} \\

\bottomrule
\end{tabular}
}

\label{tab:generalresults}
\end{table}

\begin{table}[t]
    \caption{Results on Referring Expression Comprehension (REC) task.}
    \vspace{-0.1in}
    \centering
    \tabcolsep=1mm  
    \scalebox{0.93}{
        \begin{tabular}{l|ccc|ccc|cc|c}
            \toprule
            \multirow{2}{*}{Method} & \multicolumn{3}{c}{RefCOCO} & \multicolumn{3}{c}{RefCOCO+} & \multicolumn{2}{c}{RefCOCOg} & \multirow{2}{*}{Avg}\\ 
            & val & test-A & test-B & val & test-A & test-B & val & test \\ \midrule
            GPV-2~\cite{gpv-2} & 51.59 & -- & -- & -- & -- & -- & --  & -- & -- \\
            OFA-L~\cite{wang2022ofa} & 79.96 & 83.67 & 76.39 & 68.29 & 76.00 & 61.75 & 67.57 & 67.58 & 72.65\\
            Unified-IO~\cite{unified_io} & 78.60 & -- & -- & -- & -- & -- & --  & -- & --\\
            OFASys~\cite{bai2022ofasys} & -- & 80.10 & -- & -- & -- & -- & --  & -- & --\\
            VisionLLM-H~\cite{wang2023visionllm} & -- & 86.70 & -- & -- & -- & -- & -- & --  & -- \\    
            UNITER~\cite{chen2020uniter} & 81.41 & 87.04 & 74.17 & 75.90 & 81.45 & 66.70 & 74.02 & 68.67 & 76.17 \\
            VILLA~\cite{gan2020large} & 82.39 & 87.48 & 74.84 & 76.17 & 81.54 & 66.84 & 76.18 & 76.71 & 77.77\\
            UniTAB~\cite{yang2022unitab} & 86.32 & 88.84 & 80.61 & 78.70 & 83.22 & 69.48 & 79.96 & 79.97 & 80.89 \\
            MDETR~\cite{kamath2021mdetr} & 86.75 & 89.58 & 81.41 & 79.52 & 84.09 & 70.62 & 81.64 & 80.89 & 81.81\\
            
            \midrule
            
            Shikra~\cite{shikra} & 87.01 & 90.61 & 80.24 & 81.60 & 87.36 & 72.12 & \textbf{82.27} & 82.19 & 82.92  \\ [3pt]
    
             Shikra$_{+\text{G-350K(ours)}}$ & \textbf{87.48} & \textbf{91.05} & \textbf{81.77} & \textbf{81.89} & \textbf{87.43} & \textbf{73.14} & 81.99 & \textbf{83.15} & \textbf{83.49}\\ 
    
            \bottomrule
        \end{tabular}
    }
    \label{tab:groundingresults}
\end{table}

\noindent\textbf{Evaluation on Grounding Tasks.}

\noindent\textbf{Benchmarks.} Following the baseline model Shikra~\cite{shikra}, we test the enhanced model on REC tasks. We use the test datasets RefCOCO~\cite{refcoco}, RefCOCO+~\cite{refcoco}, RefCOCOg~\cite{refcocog}.

\noindent\textbf{Main results.} Here, we adopt Shikra~\cite{shikra} as the model to evaluate the quality of Genixer-350K produced by \genixerS{}. Adhering to Shikra's published code, we incorporate our synthetic data into the training phases, maintaining consistent training iterations to ensure a fair comparison. Tab.~\ref{tab:groundingresults} shows the improvement on 7 out of 8 test datasets with a non-trivial average boost of 0.6\%. These findings imply that our pipeline can be an alternative approach to generate grounding-based instruction tuning data, which is typically challenging for manually labeling and not satisfied for prompting GPT-4V, as shown in Fig.~\ref{fig:fig1}.

\noindent\textbf{Performance on Genixer.} As illustrated in Tab.~\ref{tab:fuyuresults} and Fig.~\ref{fig:genixerprob}, \genixerL{} showcases a superior capability of generating high-quality data. It is natural for us to investigate the performance of \genixerL{}. Accordingly, we evaluate \genixerL{} and report its results in Tab.~\ref{tab:genxierresults}, which refers to the setting ZS. One can observe the minor declines in performance on the GQA, ScienceQA, and SEED, alongside a modest enhancement on VQAv2, VizWiz, and POPE. Such results are due to the exclusive training on generating instruction tuning data. Thus, for a fair comparison to investigate the capability of \genixerL{}, we proceeded to retrain the \genixerL{} with the mixture of the 665K finetuning dataset used in LLaVA1.5 and the datasets for training \genixerL{} for one epoch following the same training protocols. The outcomes of this process are presented as MixT in Tab.~\ref{tab:genxierresults}, where we witnessed significant improvements across all six benchmarks.

\begin{table}[t]
\centering
\caption{Performance of \genixerL{} on 6 representative benchmarks.}
\vspace{-0.1in}
\renewcommand\arraystretch{1.1}
\tabcolsep=1.4mm 

\begin{tabular}{p{16mm} p{10mm} cccccc}
\toprule
Method & Setting & VQAv2 & GQA & VizWiz & SQA$^\text{I}$ &POPE & SEED$^\text{I}$\\
\midrule

LLaVA1.5 & - & 78.5 & 62.0 & 50.0 & 66.8 & 85.9 & 66.2 \\
\genixerL{} & ZS & 79.7$_{+1.2}$ & 61.3$_{-0.7}$ & 50.8$_{+0.8}$ & 65.6$_{-1.2}$ & 86.0$_{+0.1}$ & 65.6$_{-0.6}$ \\
\genixerL{} & MixT & 80.2$_{+1.7}$ & 63.1$_{+1.1}$ & 54.1$_{+4.1}$ & 67.1$_{+0.3}$ & 87.5$_{+1.6}$ & 67.1$_{+0.9}$\\

\bottomrule
\end{tabular}

\label{tab:genxierresults}
\end{table}

\begin{table}[t]
    \renewcommand\arraystretch{1.1}

    \begin{minipage}[t]{0.48\linewidth}
    \centering
        \caption{The effect of Data scales on synthetic VQA-like dataset.}
        \vspace{-0.1in}
        \scalebox{0.7}{
        \begin{tabular}{lccccccc}
        \toprule
        Dataset & VQAv2 & GQA & VizWiz & SQA$^\text{I}$ & POPE & SEED$^\text{I}$\\
        \midrule
         Baseline & 78.5 & 62.0 & 50.0 & 66.8 & 85.9 & 66.2\\
         Genxier-300K & 79.0 & 62.9 & 52.7 & 68.5 & 87.1 & 65.8\\
         Genxier-610K & 79.0 & 63.1 & 53.7 & 69.2 & 87.2 & 66.2\\ 
         Genixer-915K & \textbf{79.1} & \textbf{63.1} & \textbf{53.8} & \textbf{69.7} & \textbf{87.3} & \textbf{66.6}\\ 
        \bottomrule
        \end{tabular}
    }
    \label{tab:datascale}
    \end{minipage}
    \hfill
    \begin{minipage}[t]{0.48\linewidth}
        \caption{The effect of different probability threshold $\lambda$.}
        \vspace{-0.1in}
        \scalebox{0.7}{
        \begin{tabular}{l ccccccc}
        \toprule
        Setting & Size & VQAv2 & GQA & VizWiz & SQA$^\text{I}$ & POPE & SEED$^\text{I}$\\
        \midrule
         Baseline & - & 78.5 & 62.0 & 50.0 & 66.8 & 85.9 & 66.2 \\ 
         $\lambda = 0$  & 1.4M & 79.0 & 62.9 & 53.5 & 69.6 & 87.1 & 66.2\\
         $\lambda = 0.5$ & 1.1M & 79.1 & 63.1 & 53.2 & 69.1 & 86.9 & 66.4\\
         $\lambda = 0.7$ & 0.9M & \textbf{79.1} & \textbf{63.1} & \textbf{53.8} & \textbf{69.7} & \textbf{87.3} & \textbf{66.6}\\
        \bottomrule
        \end{tabular}
    }
    \label{tab:probthreshold}
    \end{minipage}


\end{table}

\subsection{Ablation Study}
\noindent\textbf{Effect of data scale.}   Tab.~\ref{tab:datascale}  investigates the effects of the scales of our synthetic data in the pretraining stage. One can observe that a larger scale often leads to a steady performance improvement on all of the six benchmarks, showing the quality of our synthetic data. 

\noindent\textbf{Probability Range.} Tab.~\ref{tab:probthreshold} investigates the impact of the probability threshold during data filtering in Section~\ref{sec:dgandfilter} on the data quality. By varying $\lambda$, we observe that higher values of $\lambda$ often better improve performance across all six benchmarks,  even with a reduced number of selected training samples. This suggests that the quality of data is more crucial than the quantity of samples.

\subsection{Visualizations}
We sample some examples of general tasks such as Common VQA, MC VQA, and MD in Fig.~\ref{fig:commonvisual}, ~\ref{fig:mcvisual}, ~\ref{fig:mdvisual}, respectively. The examples of grounding tasks are shown in Fig.~\ref{fig:groundingvisual1} and ~\ref{fig:groundingvisual2}. These visualizations demonstrate that our \genixerL{} and \genixerS{} have the exceptional capability of generating diverse instruction tuning data. The generated VQA triplets in Fig.~\ref{fig:commonvisual} show that \genixerL{} is capable of generating diverse types of data such as color, Yes/No, and counting types. The generated samples of Fig.~\ref{fig:mcvisual} show the ability to generate multi-choice VQA data at a high-quality level. As for generating MD data, we display four examples in Fig.~\ref{fig:mdvisual}. The results demonstrate the ability to generate long semantic sentences when answering the questions. Furthermore, we also compare the REC data generation ability with GPT-4V~\cite{gpt4v} in Fig.~\ref{fig:comparegpt4v}. It indicates that GPT-4V is suboptimal for generating grounding tasks such as REC.

\subsection{User Study}

\begin{figure}
    \centering
    \includegraphics[width=0.66\linewidth]{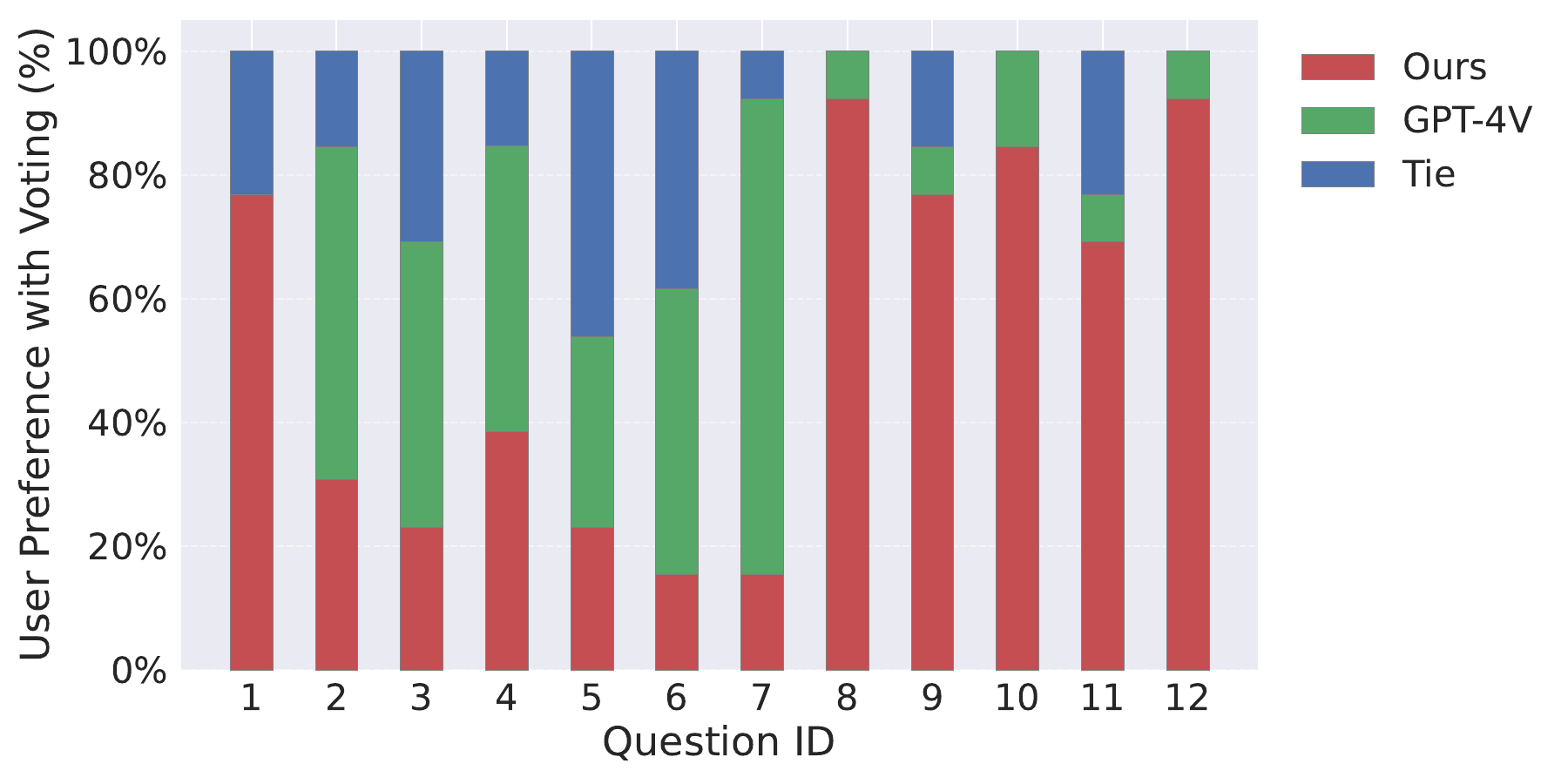}
    \vspace{-0.1in}
    \caption{User preference with voting for comparing the generated data quality between \genixer{} and GPT-4V.}
    \label{fig:userstudy}
\end{figure}

\begin{figure}
    \centering
    \includegraphics[width=\linewidth]{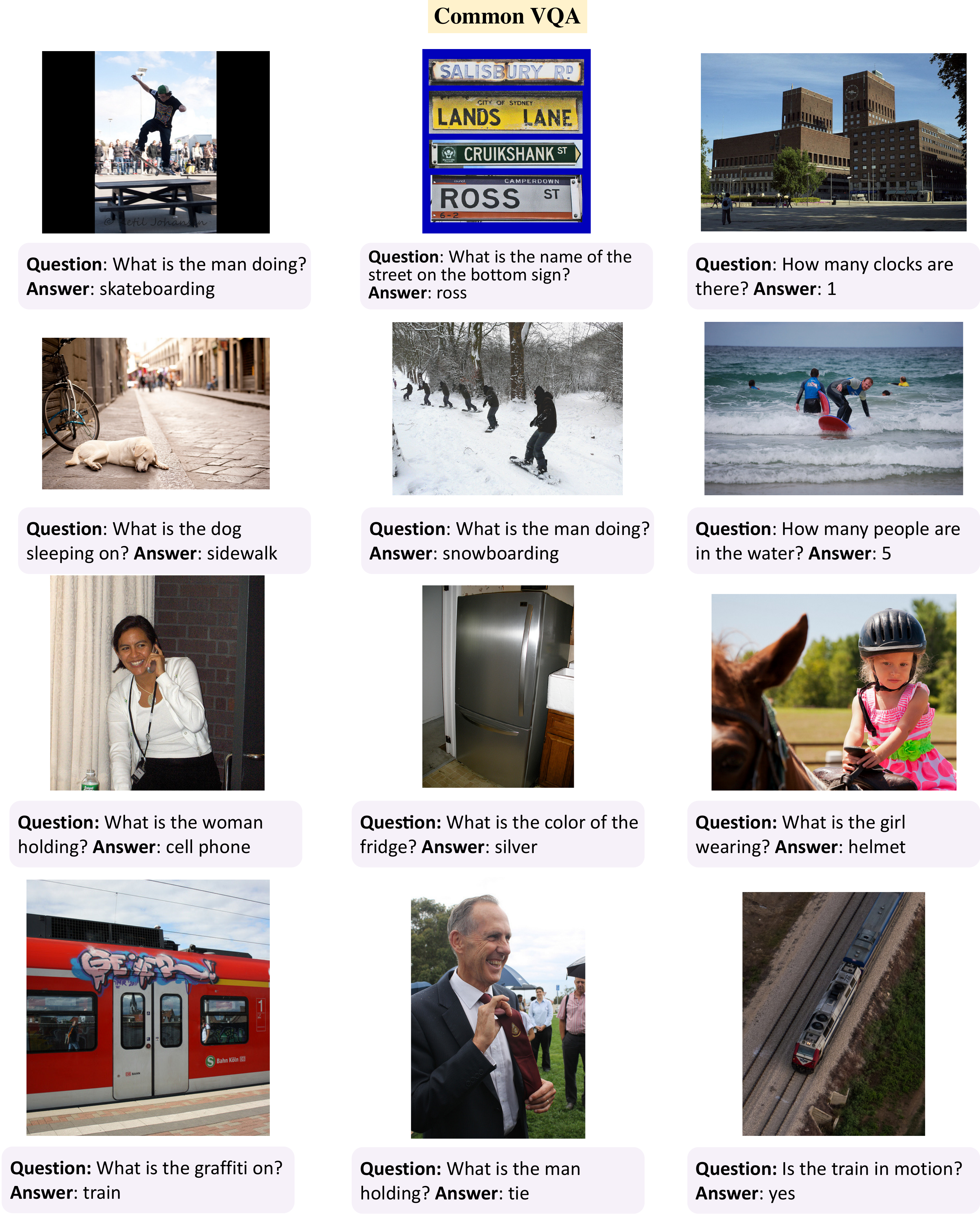}
    \caption{Some examples of Common VQA generated by \genixerL{}.}
    \label{fig:commonvisual}
\end{figure}

\begin{figure}
    \centering
    \includegraphics[width=\linewidth]{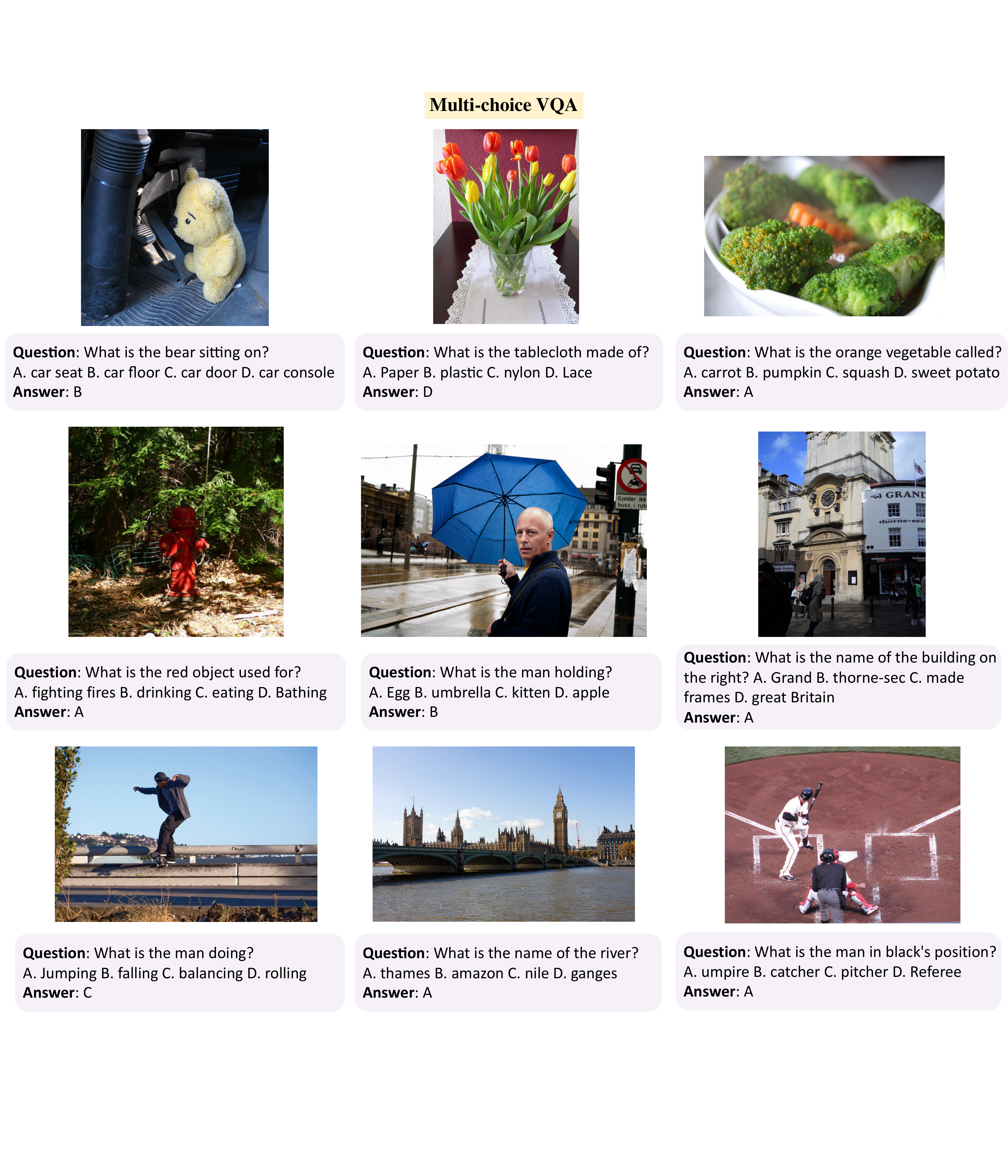}
    \caption{Some examples of Multi-choice VQA generated by \genixerL{}.}
    \label{fig:mcvisual}
\end{figure}

\begin{figure}
    \centering
    \includegraphics[width=\linewidth]{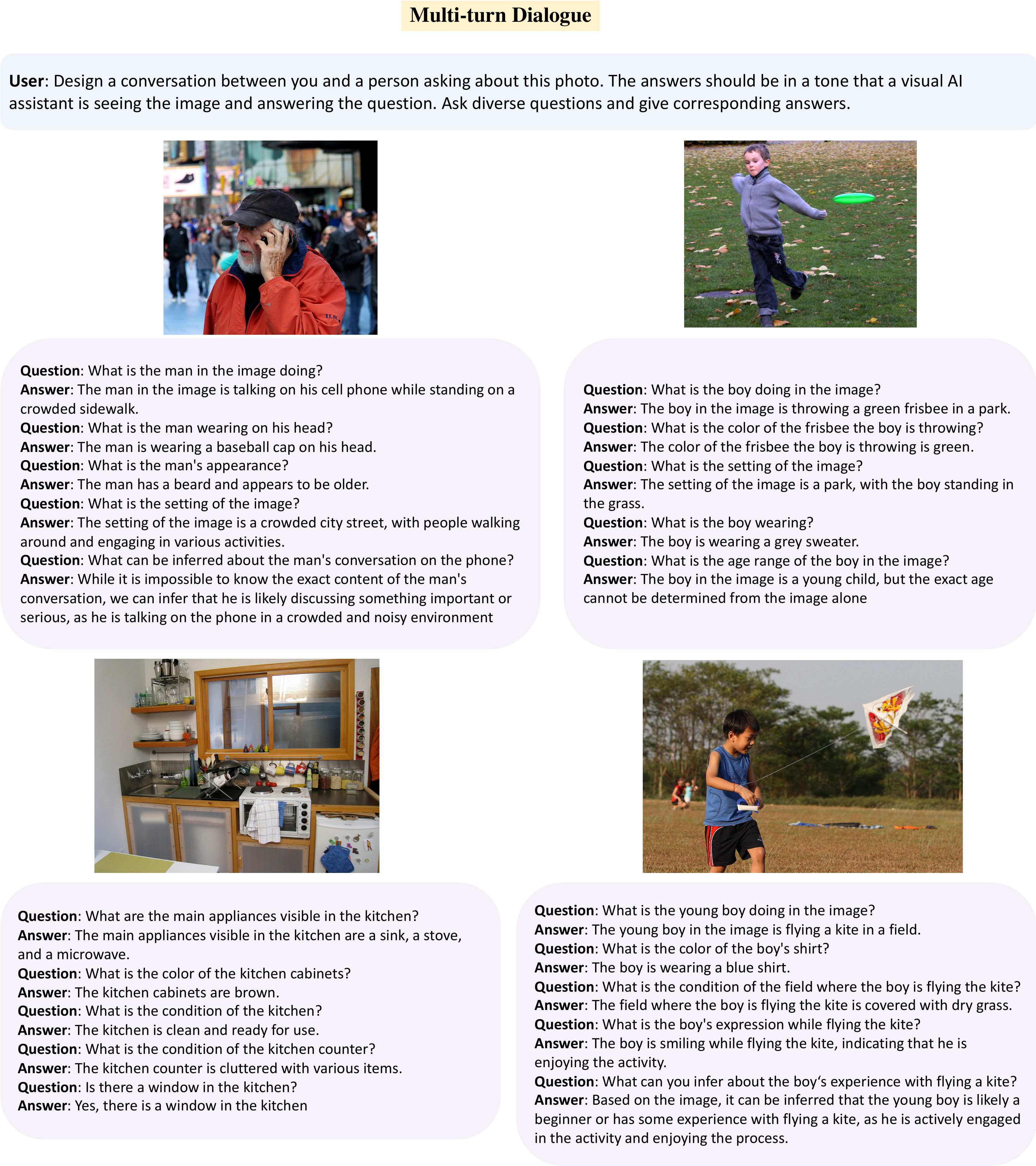}
    \caption{Some examples of Multi-turn Dialogue generated by \genixerL{}.}
    \label{fig:mdvisual}
\end{figure}

\begin{figure}
    \centering
    \includegraphics[width=\linewidth]{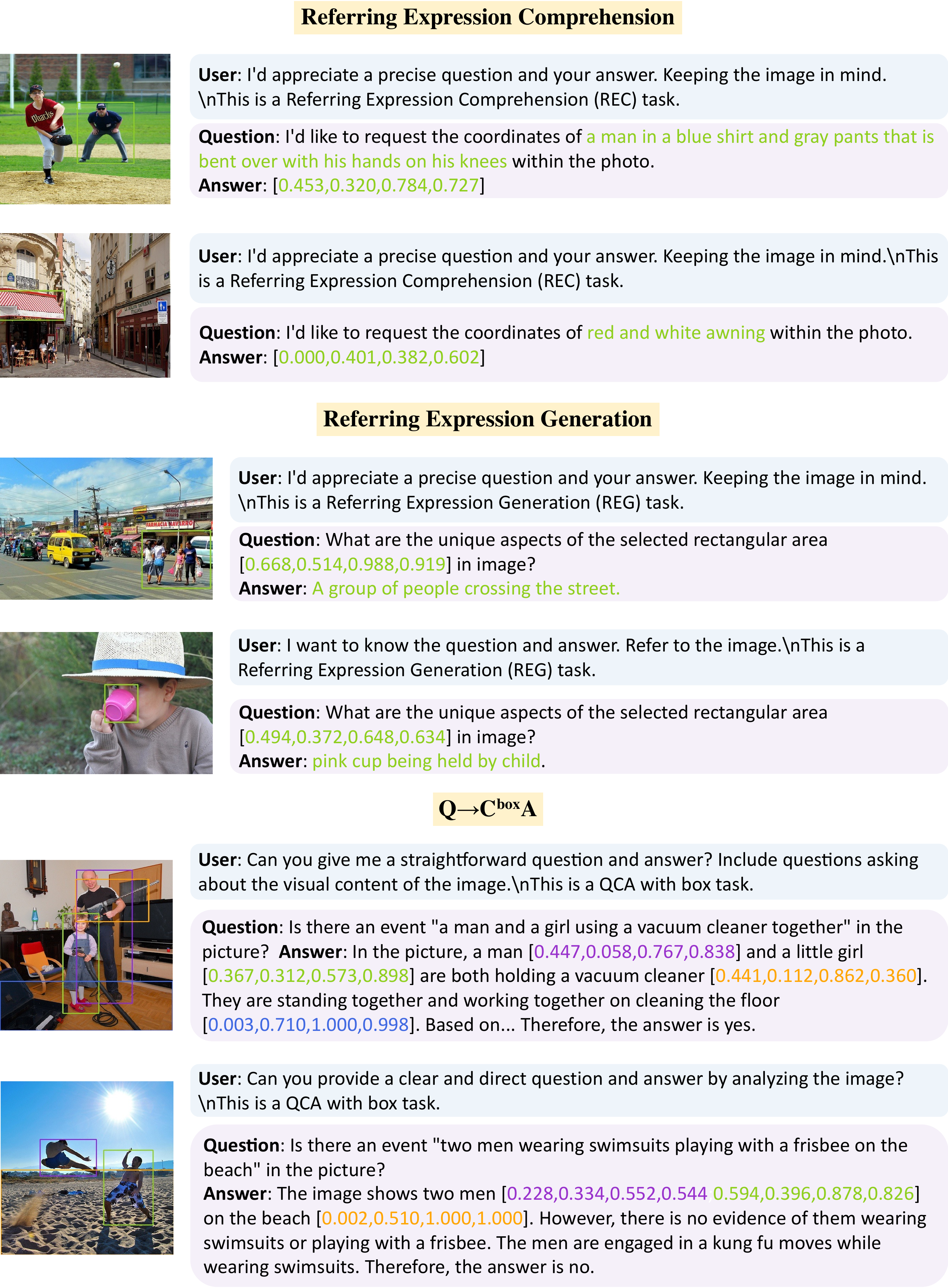}
    \caption{Some examples of grounding tasks generated by \genixerS{}.}
    \label{fig:groundingvisual1}
\end{figure}

\begin{figure}
    \centering
    \includegraphics[width=\linewidth]{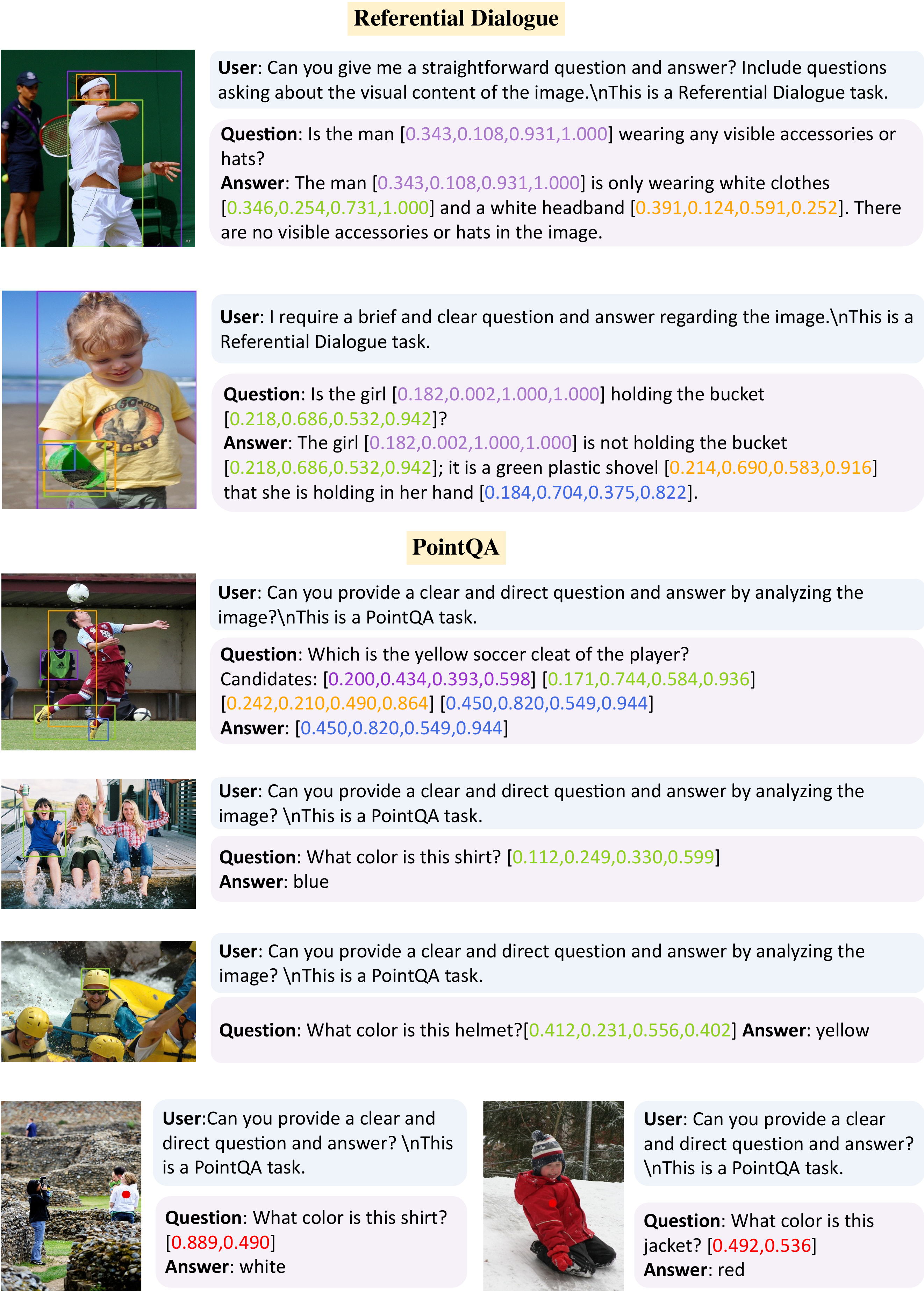}
    \caption{Some examples of grounding tasks generated by \genixerS{}.}
    \label{fig:groundingvisual2}
\end{figure}

\begin{figure}
    \centering
    \includegraphics[width=\linewidth]{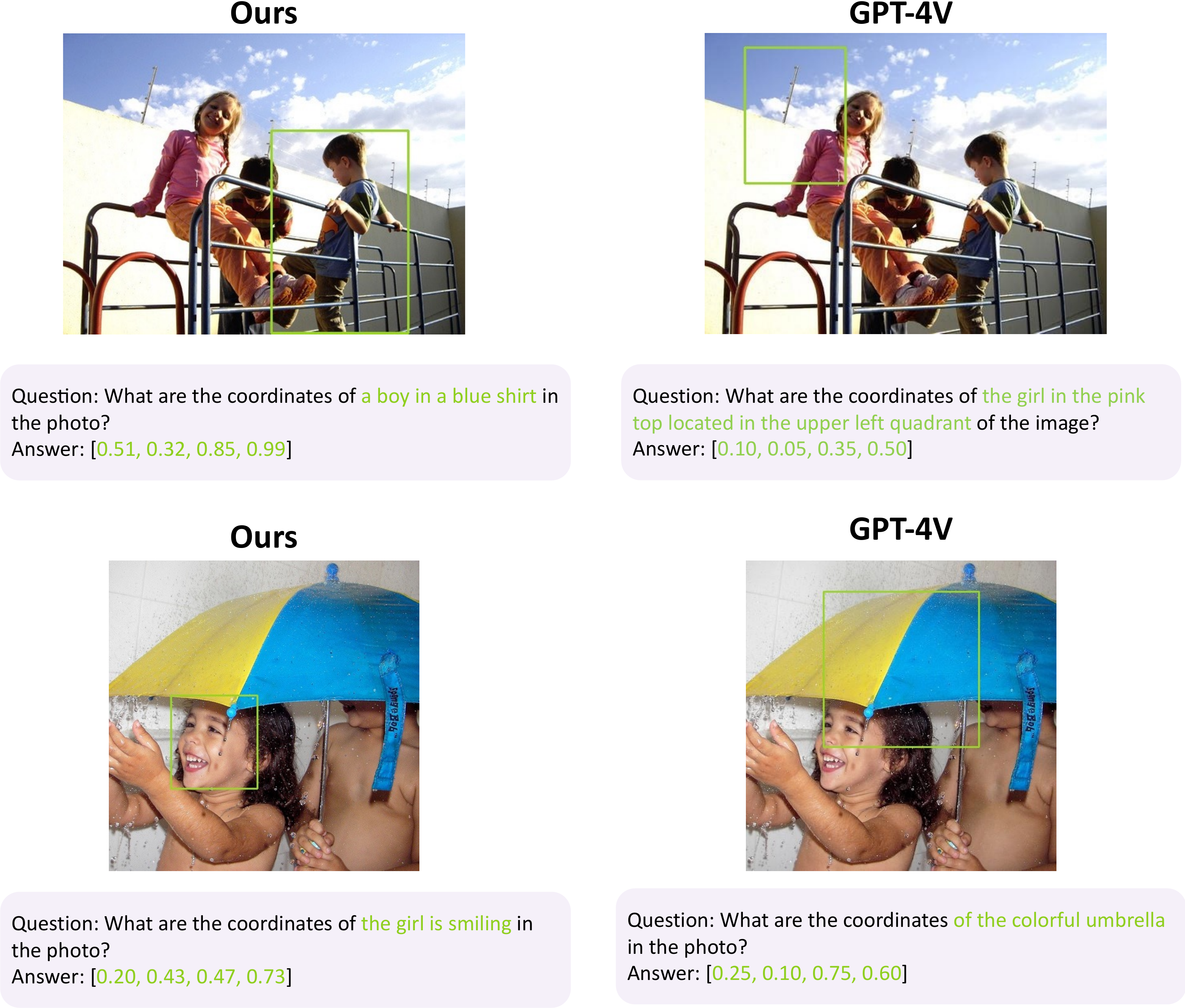}
    \caption{Comparison of generating REC-like data between \genixerS{} and GPT-4v~\cite{gpt4v}.}
    \label{fig:comparegpt4v}
\end{figure}

We conducted a user preference study to evaluate the generation quality between \genixer{} and GPT-4V. For this study, we selected 12 samples, comprising 4 Common VQA samples, 3 MC VQA samples, and 5 REC samples. Fig.~\ref{fig:userstudy} summarizes the statistical analysis of 13 valid surveys. (1) The first seven columns (excluding the first one) reveal that GPT-4V was the primary preference among users, while between 20\% to 40\% of users preferred the data generated by \genixer{}. Notably, a significant number of users selected ``Tie'' indicating that the data quality we generated is comparable to that of GPT-4V. (2) In analyzing the last five columns, our generated REC samples were more favored than those of GPT-4V, as the evidence shown in Fig.~\ref{fig:comparegpt4v}.

\section{Conclusion, Limitations, and Societal Impacts}
In this paper, we introduce a novel automatic data generation pipeline called \genixer{}, designed to efficiently and affordably produce high-quality instruction tuning data by leveraging current MLLMs. We instantiate \genixer{} into two data-generative MLLMs, \genixerL{} and \genixerS{}, tailored to generate general and grounding instruction tuning data, respectively. To ensure the quality of the generated data, we propose two data filtering frameworks: Fuyu-driven and CLIP-driven. Finally, we contribute two instruction tuning datasets, Genixer-915K and Genixer-350K, targeting Common VQA and REC. Experimental results demonstrate that both generated datasets significantly enhance LLaVA1.5 and Shikra across various multimodal benchmarks, respectively.  

\textbf{Limitations.} \textit{1) LLM Scale:} Due to computational constraints, we do not test larger LLM model, such as 13B or 34B. But we believe that our data generator could be  beneficial, since larger models are more data hungry.  \textit{2) Data Scale:} While scaling up the candidate image corpus to larger datasets like LAION-2B could enhance model capability, training costs and time constraints restrict us to do such expansions. But Tab.~\ref{tab:datascale} shows scaling can improve performance. \textit{3) Evaluation:} Despite proposing effective data filtering frameworks, evaluating complex and open-ended data types like Referential Dialogue remains challenging, leaving room for future exploration.

\textbf{Societal Impacts.} Our work addresses the challenge of generating high-quality instruction tuning data by presenting a comprehensive pipeline. It paves the way for future investigations into generating diverse multimodal data, contributing to advancements in various fields.

%
%
\bibliographystyle{splncs04}
\bibliography{main}
\end{document}